%% file: acl_latex.tex
\pdfoutput=1

\documentclass[11pt]{article}

\usepackage[]{acl}

\usepackage{times}
\usepackage{latexsym}

\usepackage[T1]{fontenc}

\usepackage[utf8]{inputenc}

\usepackage{microtype}

\usepackage{inconsolata}

\usepackage{url}
\usepackage{booktabs}
\usepackage{graphicx}
\usepackage{subcaption}
\usepackage{CJKutf8}
\usepackage{multirow}
\usepackage{xspace}
\usepackage{array}
\usepackage{amsmath}

\input{latex/macros}

\newcommand*{\affaddr}[1]{#1} 
\newcommand*{\email}[1]{\texttt{#1}}

\title{Is Translation All You Need?\\ A Study on Solving Multilingual Tasks with Large Language Models}

\author{
    Chaoqun Liu\thanks{$^{*}$Chaoqun Liu  is under the Joint PhD Program between DAMO Academy and Nanyang Technological University. }~~\affmark[12]\;
    Wenxuan Zhang\thanks{$^{\dag}$Wenxuan Zhang is the corresponding author.}~~\affmark[23]\; 
    Yiran Zhao\affmark[24]\; 
    \textbf{Anh Tuan Luu\affmark[1]\;
Lidong Bing\thanks{$^{\ddag}$Work done while at Alibaba Group.}~~\affmark[5]}\\
    \affaddr{\affmark[1]Nanyang Technological University, Singapore};\\
    \affaddr{\affmark[2]DAMO Academy, Alibaba Group, Singapore};
    \affaddr{\affmark[3]Hupan Lab, 310023, Hangzhou, China};\\
    \affaddr{\affmark[4]National University of Singapore};
    \affaddr{\affmark[5]Shanda AI Research Institute}\\
    \email{\{chaoqun.liu,saike.zwx\}@alibaba-inc.com}; lidong.bing@shanda.com
    }

\begin{document}
\maketitle
\begin{abstract}
Large language models (LLMs) have demonstrated multilingual capabilities, yet they are mostly English-centric due to the imbalanced training corpora. While prior works have leveraged this bias to enhance multilingual performance through translation, they have been largely limited to natural language processing (NLP) tasks. In this work, we extend the evaluation to real-world user queries and non-English-centric LLMs, offering a broader examination of multilingual performance. Our key contribution lies in demonstrating that while translation into English can boost the performance of English-centric LLMs on NLP tasks, it is not universally optimal. For culture-related tasks that need deep language understanding, prompting in the native language proves more effective as it better captures the nuances of culture and language. Our experiments expose varied behaviors across LLMs and tasks in the multilingual context, underscoring the need for a more comprehensive approach to multilingual evaluation. Therefore, we call for greater efforts in developing and evaluating LLMs that go beyond English-centric paradigms.\footnote{Our code is publicly available at \url{https://github.com/DAMO-NLP-SG/translation-all-you-need}.}
\end{abstract}

\input{latex/sections/introcution}

\input{latex/sections/experiments}

\input{latex/sections/related_work}

\section{Conclusion}
We have conducted a thorough evaluation of LLMs in various multilingual tasks. These tasks include traditional NLP benchmarks, real user queries, and culture-related tasks. Even though translation-based methods are simple and effective strategies to overcome the limitations inherent in English-centric LLMs, they are not optimal for all scenarios, highlighting the necessity of more comprehensive multilingual evaluation. The experiment on non-English-centric LLMs and culture-related tasks demonstrates that employing prompts in the native language emerges as a more effective approach. This method is particularly adept at capturing the subtleties and intricacies unique to each language. The challenge of the setting is that it requires LLMs to be proficient in various languages, calling for the prioritization of research and development efforts toward the creation of strong multilingual LLMs.

\section*{Limitations}
This study aims to systematically assess the effectiveness of various prompting strategies across different tasks and LLMs. Due to limitations in computing resources, it was not possible to evaluate all existing prompting strategies comprehensively. However, we endeavoured to cover the most commonly employed strategies to formulate a broad conclusion. In our evaluation of LLMs on culture-related tasks, we specifically selected two LLMs optimized for Chinese, acknowledging it as one of the most widely spoken languages globally. The dataset used, M3Exam, comprises exclusively multiple-choice questions. It is important to note this specificity as it may influence the applicability of our findings. In our evaluation, we limited our sampling to up to 500 samples for each language within the benchmarks to manage computational constraints and ensure a broad yet feasible analysis scope. Consequently, our results might not be directly comparable with other studies that evaluate performance across the entire benchmark. 
In future work, we plan to extend our evaluation to LLMs optimized for other languages and to explore benchmarks presented in various formats beyond multiple-choice questions.

\section*{Acknowledgements}

This research is supported, in part, by DSO Singapore under the research grant DSOCL23216. This research is also supported by DAMO Academy through DAMO Academy Research Intern Program and Alibaba-NTU Singapore Joint Research Institute (JRI), Nanyang Technological University, Singapore. Chaoqun Liu extends his gratitude to Interdisciplinary Graduate Programme and College of Computing and Data Science of NTU, for their support. We sincerely appreciate the valuable feedback from Hou Pong Chan (DAMO Academy, Alibaba Group).

\bibliography{anthology,custom, references}

\appendix
\input{latex/sections/appendix}

\end{document}

%% file: latex/macros.tex
\newcommand*{\affmark}[1][*]{\textsuperscript{\rm #1}}

\newcommand{\chinese}[1]{\begin{CJK*}{UTF8}{gbsn}#1\end{CJK*}}

\newcommand{\nativebasic}{\textsc{Native-Basic}\xspace}
\newcommand{\enbasic}{\textsc{EN-Basic}\xspace}
\newcommand{\nativecot}{\textsc{Native-CoT}\xspace}

\newcommand{\encot}{\textsc{EN-CoT}\xspace}
\newcommand{\xlt}{\textsc{XLT}\xspace}

\newcommand{\transg}{\textsc{Trans-Google}\xspace}
\newcommand{\transn}{\textsc{Trans-NLLB}\xspace}

\newcommand{\cothought}{\text{chain-of-thought}\xspace}



%% file: latex/sections/introcution.tex
\section{Introduction}

Large language models (LLMs) frequently demonstrate the capability to understand and generate text across multiple languages, a skill attributed to their training on vast corpora composed of texts from various languages 
\cite{openai_gpt-4_2023, shi_language_2022, muennighoff_crosslingual_2023, jiang_mistral_2023, nguyen_seallms_2023}.
However, these datasets are often disproportionately dominated by English content \cite{brown_language_2020, chowdhery_palm_2022, workshop_bloom_2023,lin_few-shot_2022}, resulting in an English-centric bias in LLMs. This imbalance can subsequently hinder the models' proficiency in other languages, often leading to suboptimal performance in non-English contexts \cite{ahuja_mega_2023, lai_chatgpt_2023, zhang_dont_2023}.

\begin{figure}
    \centering
    \includegraphics[width=\linewidth]{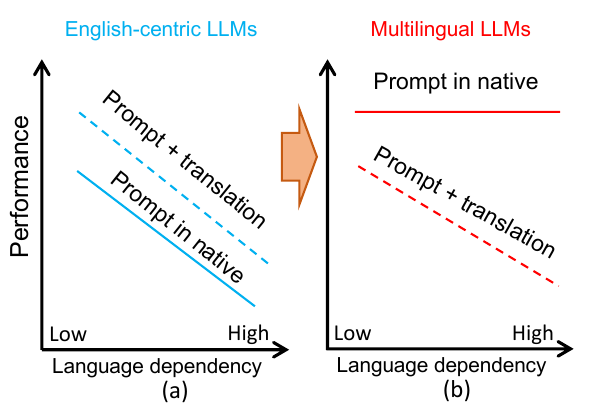}
    \caption{Illustration of two types of LLMs on tasks with varying language dependencies. "English-centric LLMs" refers to LLMs trained mainly in English corpora. "Multilingual LLMs" refers to ideal LLMs equally capable in all languages. }
    \label{fig:overview}
\end{figure}

\begin{figure*}[t]
    \centering
    \includegraphics[width=\linewidth]{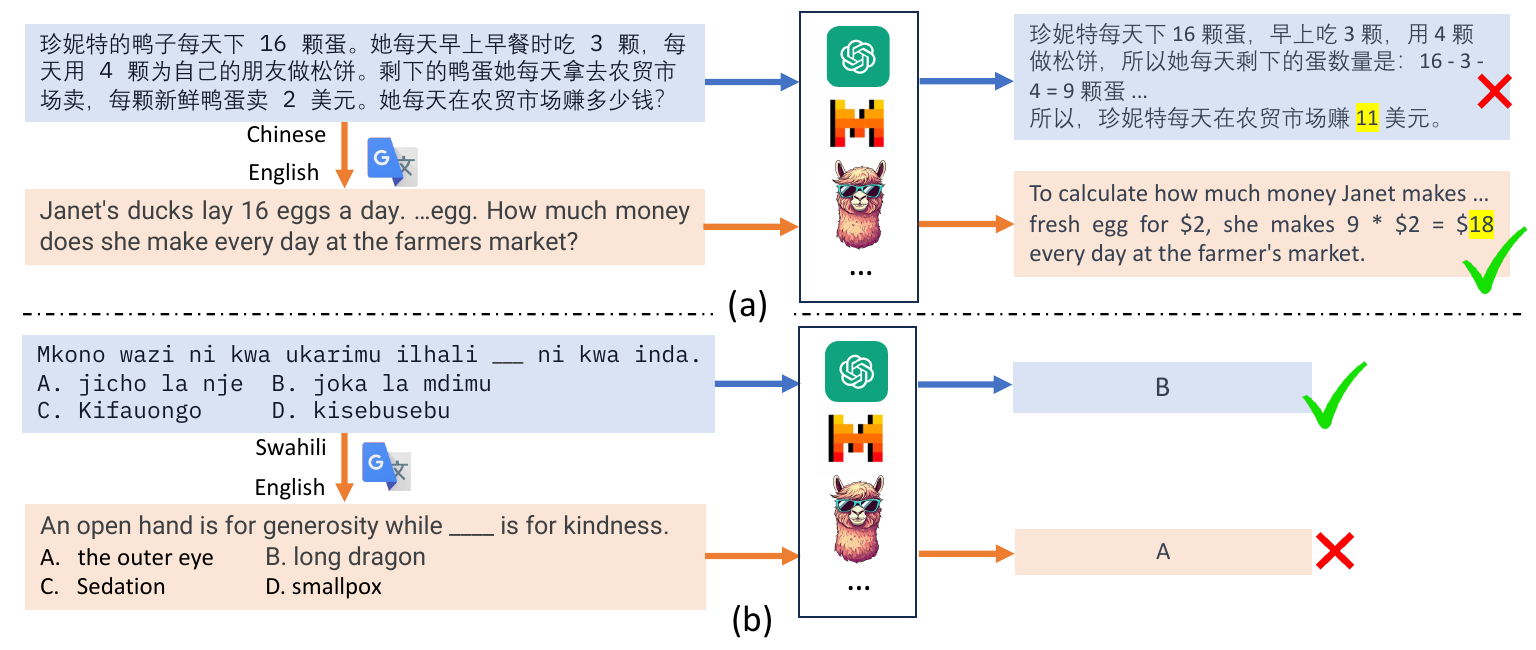}
    \caption{Examples illustrating how translation can both improve (a) and degrade (b) the performance of LLMs. The Chinese example is from MGSM \cite{shi_language_2022} and the Swahili example is from M3Exam \cite{zhang_m3exam_2023}. Translation is beneficial when the questions are semantically equivalent across languages. However, for questions that demand deep cultural knowledge, translation can hinder the ability to answer accurately.
    }
    \label{fig:illustration}
\end{figure*}

To enhance performances in multilingual natural language processing (NLP) tasks with English-centric language models,
translating training or test data into English has proven an effective strategy \cite{conneau_xnli_2018,ponti_xcopa_2020,artetxe_revisiting_2023,moghe_extrinsic_2023,bareis_english_2024}. Recent investigations have expanded this idea by incorporating translation, either implicitly or explicitly, into the intermediate stages of prompting LLMs \cite{huang_not_2023,qin_cross-lingual_2023,etxaniz_multilingual_2023} for multilingual NLP tasks. For example, \citealp{shi_language_2022} demonstrates that translating test questions into English enhances performance on multilingual reasoning tasks, as illustrated in Figure \ref{fig:illustration}(a). Similarly, \citealp{huang_not_2023} and \citealp{etxaniz_multilingual_2023} have shown that prompting LLMs to first translate or comprehend questions in English, then solve them step by step, improves performance.

Despite these advancements, methodologies in various studies differ significantly, and the impact of translation on multilingual task performance remains underexplored. Furthermore, these studies focus on specific NLP tasks and English-centric LLMs, but did not study real-world user queries in various languages. 
This gap highlights a need for more nuanced research into the effectiveness of translation techniques across multilingual contexts. As shown in Figure \ref{fig:overview}, we hypothesize that English-centric LLMs generally perform better with English translations of prompts, while "Multilingual LLMs" excel with native prompts, particularly for tasks highly dependent on language.

To address the limitations of existing empirical studies, we perform an in-depth analysis of the utility of translation with large language models for various scenarios. Firstly, we compare translating multilingual tasks into English, with an optional step of translating responses back into the original languages (i.e., the ``translate-test'' method), 
against several baselines on multilingual NLP tasks. Secondly, we extend the evaluation to real user queries, which are more likely to contain knowledge related to culture and language. Thirdly, we broaden the scope of LLM evaluations to include non-English-centric models to explore how they differ in behavior from English-centric LLMs. 
To the best of our knowledge, \textit{this is the first work to analyze the impacts of translating real user queries on multilingual LLMs}.

Our results demonstrate that simply translating queries into English can already achieve the best results in multiple NLP task categories. 
For real user queries, the effect of translation depends on the languages and the LLMs. When working with advanced LLMs and certain languages, employing prompts in native languages appears to be the more effective strategy. 
In addition, the non-English-centric LLMs also behave differently from English-centric LLMs, where prompts in the native languages yield superior results by capturing the nuances related to culture and language. 

The main contributions of this work are:
\begin{itemize}
    \item We conduct a comprehensive comparison of multilingual prompting strategies in NLP tasks, finding that translation remains a strong baseline even for LLMs, and identifying factors impacting multilingual performance.
    \item We expand multilingual evaluation to include actual user queries and and non-English-centric LLMs, addressing the limitations of previous studies.
    \item We expose critical gaps in current multilingual evaluations, underscoring the need for more comprehensive benchmarks and a broader range of LLMs.
\end{itemize}

%% file: latex/sections/experiments.tex
\section{Translation for NLP Tasks}\label{sec:ex_nlp}
This section explores various prompting strategies across multiple languages and LLMs, covering a wide range of NLP tasks. This helps us understand how different prompting methods and other factors affect task performance.

\subsection{Experiment Setup}
\subsubsection{Tasks} \label{sec:nlp_tasks}
We conduct assessments on six benchmarks covering reasoning, understanding, and generation tasks that encapsulate various abilities of LLMs: \textbf{MGSM} \cite{shi_language_2022}, \textbf{XCOPA} \cite{ponti_xcopa_2020}, \textbf{XNLI} \cite{conneau_xnli_2018}, \textbf{PAWS-X} \citep{yang_paws-x_2019}, \textbf{MKQA} \citep{longpre_mkqa_2021} and \textbf{XL-Sum} \cite{hasan_xl-sum_2021}. Following  \citealp{huang_not_2023}, we choose a subset of 9 languages for MKQA and 5 languages for XL-Sum. For evaluation metrics across our study, we employ the token overlap F1 score specifically for the MKQA dataset, the ROUGE-1 score for assessing XL-Sum, and accuracy as the standard metric for all other benchmarks. More details of the benchmarks can be found in Appendix \ref{Appendix:NLP}.

These tasks cover a wide array of 24 diverse languages, including German (de), Russian (ru), French (fr), Chinese Simplified (zh), Spanish (es), Japanese (ja), Italian (it), Vietnamese (vi), Turkish (tr), Indonesian (id), Swahili (sw), Arabic (ar), Korean (ko), Greek (el), Thai (th), Bulgarian (bg), Hindi (hi), Estonian (et), Bengali (bn), Tamil (ta), Urdu (ur), Telugu (te), Haitian Creole (ht), and Southern Quechua (qu). We categorize languages larger than 1\% frequency in Common Crawl\footnote{\url{https://commoncrawl.github.io/cc-crawl-statistics/plots/languages}} as high-resource languages (i.e., de, ru, fr, zh, es, ja, it and vi), and the rest as low-resource languages.
We exclude English since we want to evaluate the efficient prompting strategy for non-English tasks.

For each task, we sample 500 examples from the test set per language or use the entire test set if there are fewer than 500 examples. For generation tasks like MKQA and XL-Sum, answers will be translated back to the original language if the prompting strategy uses a translator.

\input{latex/tables/main_results_high_low_two_models}

\subsubsection{Models}
We mainly conduct experiments on the following two LLMs, consisting of one closed-source language model and one open-source language model:

\paragraph{ChatGPT} This is the most capable and cost-effective model in the GPT-3.5\footnote{\url{https://platform.openai.com/docs/models/gpt-3-5}} family optimized for chat. We chose the latest version (gpt-3.5-turbo-1106) for the experiment. 
\paragraph{Llama-2-70B-Chat} This is the largest chat models in Llama-2 family \cite{touvron_llama_2023}. Due to computational resource limitations, we use the AWQ \cite{lin2023awq} version for evaluation.

We also conducted experiments on some other models, including Mistral-7B-Instruct (v0.2) \cite{jiang_mistral_2023}, Llama-2-13B-chat \cite{touvron_llama_2023} and bloomz-7b1 \cite{muennighoff_crosslingual_2023}. More details are shown in Appendix \ref{Appendix:NLP}.

\subsubsection{Prompting Strategies}\label{sec:strategy}
We assess experimental strategies based on language of instruction, chain-of-thought reasoning, and translation tools, using a zero-shot approach as the selected models are fine-tuned for instruction-following.

\paragraph{Basic prompt with native instructions (\nativebasic)} 
The questions are posed directly without using prompting strategies like chain-of-thought. Both the query and instructions are presented in their original language.

\paragraph{Basic prompt with English instructions (\enbasic)} Compared with \nativebasic, \enbasic \ 
instructs LLMs with English but the query information is in the original language.

\paragraph{Native chain-of-thought (\nativecot)} In \nativecot, we ask the question in the native language and ask the model to reason with the native language with the instruction "\textit{Let's think step by step.}" translated into that language.

\paragraph{English chain-of-thought (\encot)} We pose the question in the native language but instruct the model to reason in English with the instruction "\textit{Let's think step by step in English}".

\paragraph{Cross-lingual-thought (\xlt)} \xlt \ \cite{huang_not_2023} is a state-of-the-art prompting method to handle multilingual NLP tasks. It prompts LLMs to translate the question into English and solve the problem step-by-step in English.

\paragraph{Translate to English with Google Translate (\transg)} It uses Google Translate API to translate the original questions into English and then solve the problem step by step.

\paragraph{Translate to English with NLLB models (\transn)} Instead of using commercial translators, we use an open-source model, namely NLLB \cite{nllb_team_no_2022}. Specifically, we chose \texttt{nllb-200-3.3B} to do the translation.

The examples for each strategy are shown in Table \ref{tab:prompt_example} and the templates for \enbasic are shown in Table \ref{tab:basic-prompt} in the Appendix.
In addition to the prompting strategies, an output constraint is also included in the template to facilitate answer extraction.  When the output format may deviate from the instructions, we utilize "\textit{Therefore, the answer <constraint> is}" in appropriate languages in the second round to retrieve the ultimate answer.

\subsection{Main Results} \label{NLP_results}
The main results are shown in Table \ref{tab:main_NLP_HL_two_models}. 
We notice that \transg, despite simple, demonstrates the highest overall performance across various models and tasks. 
While it may not always achieve top performance, 
it consistently delivers commendable results for both high and low-resource languages.
Besides this, we can have the following observations: 
1) Utilizing English instructions generally enhances performance across various tasks, regardless of the integration of \cothought. This finding aligns with those reported by \citealp{lai_chatgpt_2023}.
2) \cothought is quite helpful for strong LLMs like ChatGPT and reasoning tasks like MGSM. For weaker models and tasks that can be answered directly, the basic prompt may be a better option. 
3) On average, \encot underperforms compared to \transg for both high and low-resource languages. While \encot surpasses \transn in high-resource languages, it falls short in low-resource ones. We hypothesize that this discrepancy arises because LLMs excel in high-resource languages but need external translation systems to handle low-resource languages effectively. 

These findings are also applicable to smaller models, such as Mistral-7B-Instruct, as demonstrated in Table~\ref{tab:main_NLP_HL} in the Appendix. This suggests that the observations generalize well across different model types and sizes. Further results and discussions are provided in Appendix \ref{appendix:results}.

\subsection{Analysis and Discussions}
To investigate the impact of different factors on performance across various languages, we conduct a series of experiments and analyses using the MGSM benchmark. 

\paragraph{Is there a relationship between task performance and translation quality?}
In addition to external translation systems, we can use LLMs to translate the questions. Although \xlt includes translation, it is integrated into the solutions. Therefore, we examine the self-translate approach \cite{etxaniz_multilingual_2023}, translating in a zero-shot manner with the prompt template shown in Appendix \ref{app:nlp_prompt}.
Then we prompt LLMs with the translated question the same as \transg and \transn. The results are shown in Table \ref{tab:result_mgsm_self-translate} in the Appendix.

\begin{figure}[t]
    \centering
    \includegraphics[width=\linewidth]{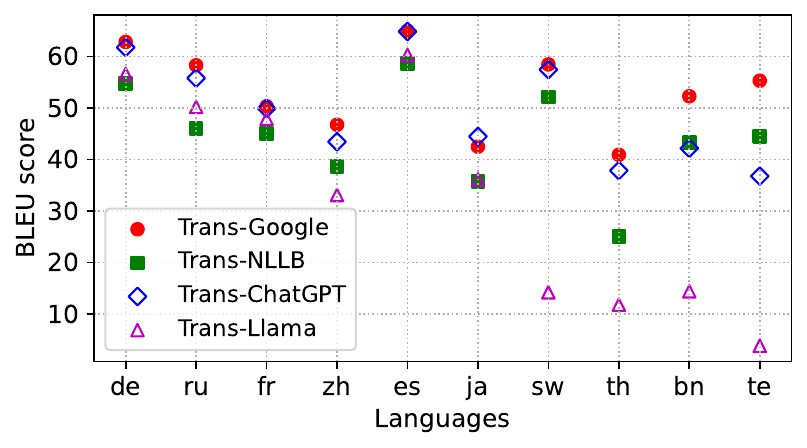}
    \caption{BLEU scores for translating MGSM questions with different translation systems.}
    \label{fig:bleu_score}
\end{figure}

We use the English subset of MGSM as the reference translation and evaluate translation quality using the SacreBLEU score \cite{papineni_bleu_2002,post_call_2018}. The results, shown in Figure \ref{fig:bleu_score}, indicate that Google Translate achieves the highest quality for all languages except Japanese. Translations by ChatGPT (Trans-ChatGPT) and Llama-2-70B-Chat (Trans-Llama) outperform \transn for high-resource languages but not for some low-resource languages. 

\begin{figure}[t]
    \centering
    \includegraphics[width=\linewidth]{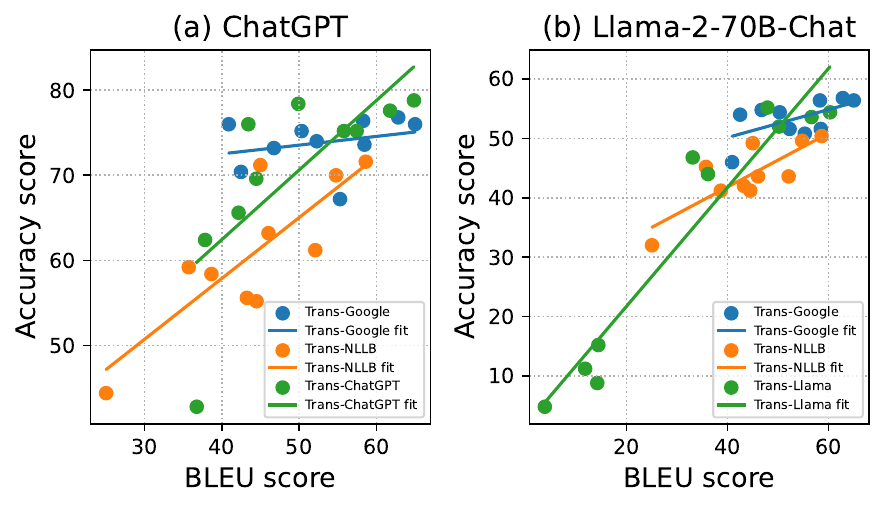}
    \caption{Corrections between BLEU scores of translation and MGSM accuracy for the three prompting techniques: \transg, \transn and self-translate. Each dot in the figure represents the performance of one model on one language.}
    \label{fig:acc_vs_bleu}
\end{figure}

To analyze the impact of translation quality on final performance, we plot the correlation between accuracy scores and BLEU scores for each language in Figure \ref{fig:acc_vs_bleu}. The results show that higher translation quality (BLEU scores) generally leads to better task performance, highlighting the importance of an effective translation system.

\paragraph{Does language distance between English and target language affect the performances?}
Table \ref{tab:main_NLP_HL_two_models} shows that the LLMs perform better for high-resource languages than low-resource languages on average. We hypothesize that language distance, besides language frequency, is crucial for English-centric LLMs. To verify this, we calculate the correlation between MGSM accuracy and the language distances between the target languages and English. Following \citealp{philippy_identifying_2023}, we examine five types of distances, including the syntactic (SYN), geographic (GEO), inventory (INV), genetic (GEN), and phonological (PHON) distances extracted using \texttt{lang2vec} \cite{littell2017uriel}.
As shown in Table \ref{tab:distance}, MGSM accuracy significantly correlates with syntactic distance but not with other types of distances. The negative values indicate that languages with a larger syntactic distance from English tend to perform worse. 

\input{latex/tables/analysis_distance}

\section{Translation for Real User Queries}
\label{sec:real_query}

\begin{figure}[t]
    \centering
    \begin{subfigure}[b]{0.45\textwidth}
        \includegraphics[width=\textwidth]{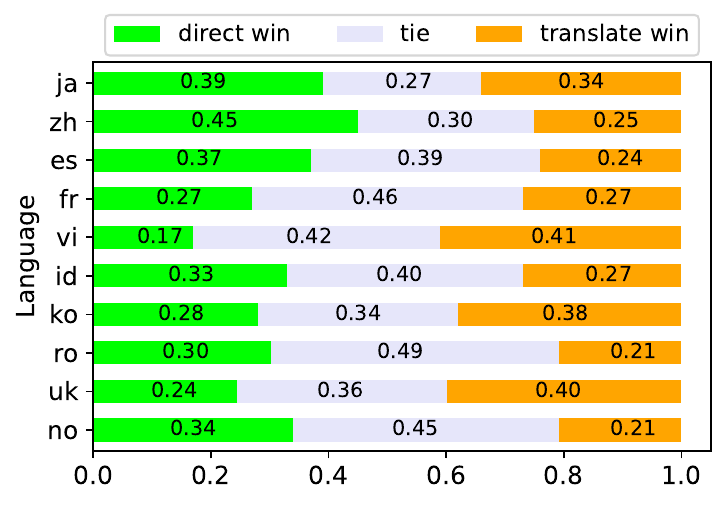}
        \caption{Win rate with ChatGPT}
        \label{fig:sub1}
    \end{subfigure}
    \hfill 
    \begin{subfigure}[b]{0.45\textwidth}
        \includegraphics[width=\textwidth]{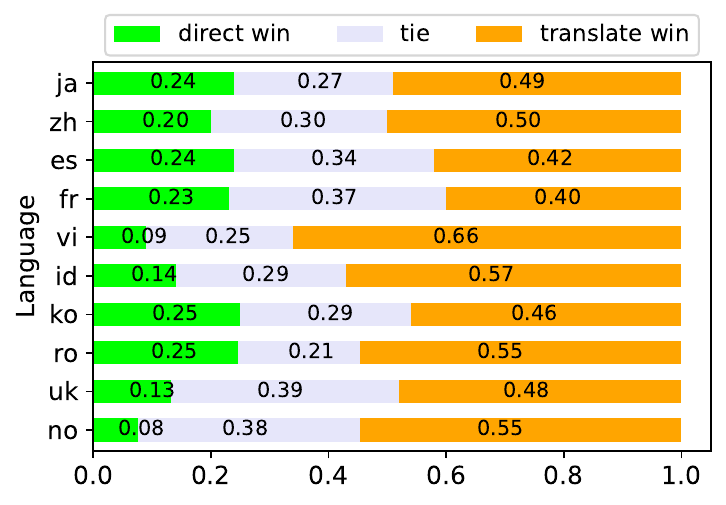}
        \caption{Win rate with Llama-2-70B-Chat}
        \label{fig:sub2}
    \end{subfigure}
    \caption{Win rate comparison for each language using ChatGPT and Llama-2-70B-Chat.}
    \label{fig:results_shareGPT}
\end{figure}

NLP tasks typically focus on specific linguistic aspects, which may not fully encapsulate the breadth and complexity of real-world user queries which cover diverse topics and require nuanced comprehension. Moreover, these benchmarks are often constructed by translating from the English data \cite{shi_language_2022,ponti_xcopa_2020,conneau_xnli_2018,yang_paws-x_2019,hasan_xl-sum_2021}. This approach leads to datasets that are not truly challenging, as they miss the rich culture-specific elements crucial for truly nuanced language understanding for different languages.
To assess the impact of translation on real-world queries, we extract user requests from ShareGPT\footnote{\url{https://sharegpt.com/}}, a website to share real conversations with ChatGPT.

\subsection{Experiment Setup} \label{sec:real_query_setup}
 We selected 10 languages, ranging from high to low resource, and randomly sampled 100 requests for each language. However, for Romanian (ro), Ukrainian (uk), and Norwegian (no), we sampled 53, 98, and 53 requests respectively, due to the limited number of samples available from the source dataset. Since the queries can be in various formats, we only compare two prompting strategies: 1) original queries; and 2) translated queries with Google Translate API. For the second option, we translate the output back to the original language for consistency. To evaluate the quality of the responses, we use GPT-4o\footnote{\url{https://openai.com/index/hello-gpt-4o/}}(\texttt{gpt-4o-2024-05-13}) as the judge. The prompt for the judge is shown in Figure \ref{fig:judge} in the Appendix, which is adapted from \cite{zheng_judging_2023}. With this prompt, each response will get a score from 1 to 10. 

\begin{figure*}[ht]
    \centering
    \begin{subfigure}[b]{0.495\textwidth}
        \includegraphics[width=\textwidth]{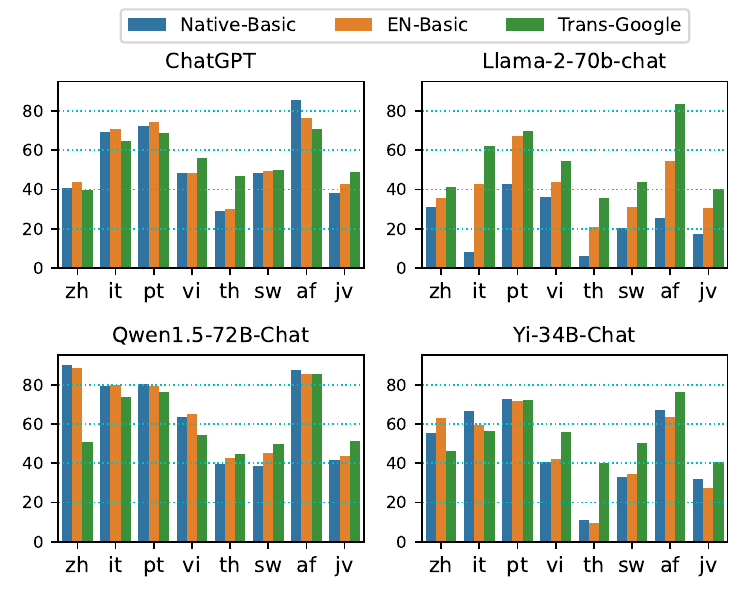}
        \caption{}
        \label{fig:M3sub1}
    \end{subfigure}
    \begin{subfigure}[b]{0.495\textwidth}
        \includegraphics[width=\textwidth]{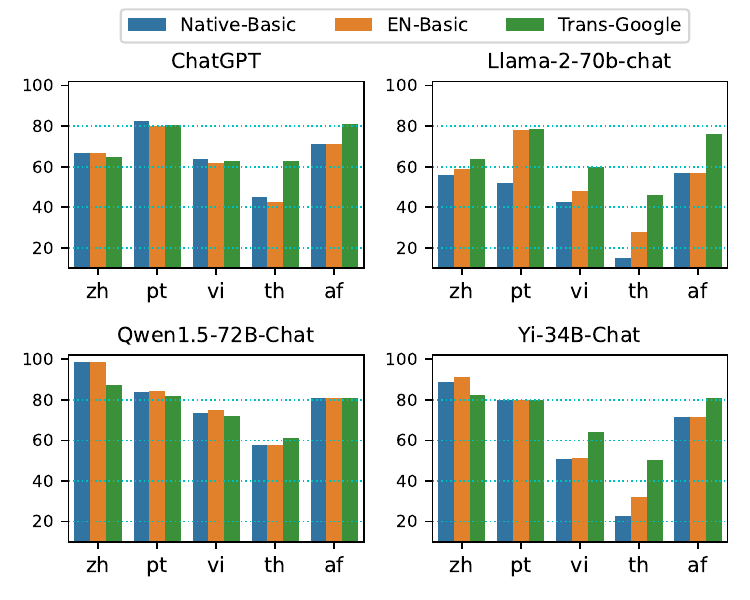}
        \caption{}
        \label{fig:M3sub2}
    \end{subfigure}
    \caption{Accuracies of four LLMs on M3Exam (a) \texttt{language} and (b) \texttt{social science} subject categories. In M3Exam, not all subjects are available in every language, causing a difference in language coverage between the two subjects.}
    \label{fig:M3Exam}
\end{figure*}

\subsection{Main Results} \label{sec:real_query_results}
We compared the scores of two response sets from the same model, calculating the win rate for each language. The results are shown in Figure \ref{fig:results_shareGPT}, leading to the following observations:
1) ChatGPT's performance varies across languages. For high-resource languages like Japanese, Chinese, and Spanish, original queries have a higher win rate. In contrast, for low-resource languages, the effectiveness of translation can be either better or worse, depending on the specific languages involved.
2) For Llama-2-70B-Chat, translation has a higher win rate for all languages, reflecting its English-centric nature. Despite potential information loss, the improved understanding after translation still enhances performance.

Llama-2-70B-Chat and ChatGPT exhibit distinct behaviors, reflecting their inherent differences. Llama-2-70B-Chat, being English-centric, performs better with translated inputs. Conversely, ChatGPT shows certain characteristics of a ``Multilingual LLM”, as shown in Figure \ref{fig:overview}(b), mainly for high-resource languages, indicating the potential for improvement in true multilingual processing.

To determine if answering user queries requires local cultural knowledge, we used GPT-4o with a specially crafted prompt to analyze queries in multiple languages (Figure \ref{fig:prompt_culture} in the Appendix). 
Results in Table \ref{tab:ratio_culture} in the Appendix show that 30\% to 74\% of queries per language require cultural knowledge, highlighting the rich cultural elements in the data. Further analysis of the ShareGPT subsets requiring local cultural knowledge is in Appendix \ref{app:shareGPT}. We also conduct additional experiments, detailed in Appendix \ref{sec:real_query_more_judges}, to verify that advanced LLMs can reliably assess the quality of responses.

\subsection{Analysis and Discussions}

Based on the previous results,
ChatGPT and Llama-2-70B-chat both tend to be English-centric but ChatGPT demonstrates certain behaviors of a "Multilingual LLM". 
Consequently, we broaden our analysis to include non-English-centric LLMs and assess their performance across various tasks.

\paragraph{How do non-English-centric LLMs perform on culture-related tasks?}

To investigate the behaviors of different LLMs on culture-related tasks, we select another two LLMs: Qwen1.5-72B-Chat \cite{qwen} and Yi-34B-Chat \cite{ai_yi_2024}, which are not English-centric. These two open-source models demonstrate strong capabilities in both English and Chinese. 
Therefore, we can check whether they demonstrate multilingual behaviors in Chinese, as illustrated in Figure \ref{fig:overview}(b).

For the evaluation dataset, we choose M3Exam \cite{zhang_m3exam_2023}, as the questions are real-world natural data from different languages instead of translating from English and require strong multilingual proficiency and cultural knowledge to perform well. For example, the question about a Swahili proverb in Figure \ref{fig:illustration}(b) requires local knowledge to answer correctly. We select the \texttt{language} and \texttt{social science} subject categories, which likely contain more native cultural knowledge, and evaluate up to 500 samples per language.

Based on the results shown in Figure \ref{fig:M3Exam}, we have the following observations: 1) For ChatGPT, translation may not always result in improved performance. This observation aligns with the conclusions in the study by \citealp{zhang_m3exam_2023}. The effectiveness of translation largely depends on whether translation errors outweigh any potential gains in better comprehension. 2) Translation helps Llama-2-70B-chat in all the languages, suggesting that the model's underperformance is due to poor language understanding rather than limitations of cultural knowledge.  3) Qwen1.5-72B-Chat and Yi-34B-Chat excel in Chinese proficiency. The translation hurts Chinese performance, highlighting the significant influence of translationese on comprehension. Despite this, it may boost performance in other languages, notably for  Yi-34B-Chat, indicating that they are far from ideal multilingual LLMs.

\paragraph{How do non-English-centric LLMs perform on NLP tasks?}
\input{latex/tables/result_bilingual_NLP}
As shown in Figure \ref{fig:illustration}(b), for an ideal multilingual LLM, prompting in native languages should still have advantages over translation if the tasks are less dependent on languages. To test the hypothesis, we evaluate Qwen1.5-72B-Chat and Yi-34B-Chat on the NLP tasks as discussed in Section \ref{sec:nlp_tasks}. We only evaluate them in Chinese since the two models are optimized for this language. 

The results are displayed in Table \ref{tab:bilingual_NLP}. \transg remains competitive among various prompting strategies, achieving the best average scores for Yi-34B-Chat, which surpasses our expectations. The possible reason could be that while both models are optimized for Chinese, their performance in Chinese still lags behind their proficiency in English.
Nevertheless, We have the following special observations for the two models. 1) For Qwen1.5-72B-Chat, the best strategy is \encot instead of \transg. We hypothesize that this prompting strategy utilizes the model's bilingual abilities and simultaneously avoids translationese. 2) Both LLMs perform better with \nativebasic for the XL-Sum dataset. We hypothesize that the dataset is more language-dependent than other tasks as it is created by considering the local context instead of simply translating from the English version \cite{hasan_xl-sum_2021}. 3) The translation benefits are less pronounced than those of ChatGPT and Llama-2-70B-Chat. For example, the gap between \transg and \nativebasic on MGSM(Chinese) for the two models are 2.8\% and 8\%. The values for ChatGPT and Llama-2-70b-Chat are 37.2\% and 16\%, respectively, which are significantly larger. 

\begin{figure}[t]
    \centering
    \begin{subfigure}[b]{0.45\textwidth}
        \includegraphics[width=\textwidth]{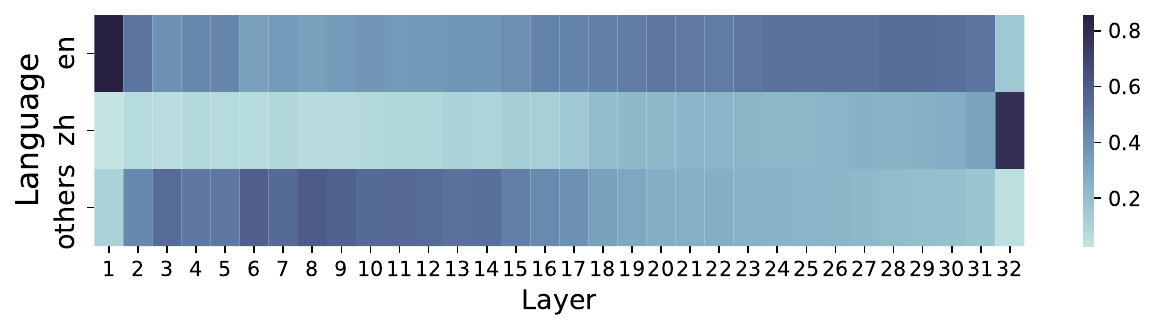}
        \caption{Llama-2-7B-Chat}
        \label{fig:layer1}
    \end{subfigure}
    \hfill 
    \hfill
    \begin{subfigure}[b]{0.45\textwidth}
        \includegraphics[width=\textwidth]{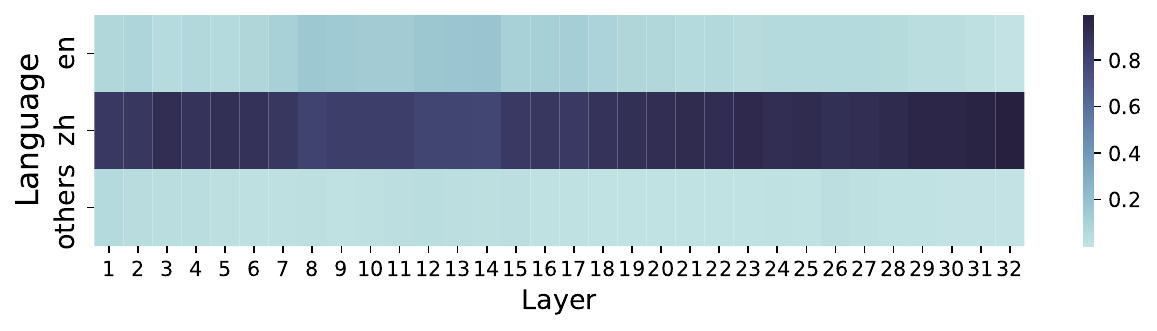}
        \caption{Qwen1.5-7B-Chat}
        \label{fig:layer2}
    \end{subfigure}
    \caption{Layerwise language distribution for (a) Llama-2-7b-chat and (b) Qwen1.5-7B-Chat with Chinese prompts.}
    \label{fig:layerwise_lang}
\end{figure}

\paragraph{How do different LLMs handle multilingual prompts?}
To further understand the differences between English-centric LLMs and non-English-centric LLMs, we analyze the layerwise language distribution for Llama-2-7B-Chat and Qwen1.5-7B-Chat, using the method proposed by \citealp{zhao_how_2024}. We decode the embedding after each layer and identify each token into different languages with CLD3\footnote{\url{https://github.com/google/cld3}}. As shown in Figure \ref{fig:layerwise_lang}, the two LLMs process Chinese prompts differently. While the hidden representations of Qwen1.5-7B-Chat are mainly in Chinese, those of Llama-2-7B-Chat are in various other languages.
We hypothesize that processing the information in native without conversion avoids the information loss, making it more suitable for processing multilingual tasks. In addition, we examine the layerwise language distribution in larger models, specifically Llama-2-70B-Chat and Qwen1.5-72B-Chat, as shown in Figure \ref{fig:layerwise_lang_2} within Appendix \ref{sec:appen_layer}.

%% file: latex/tables/main_results_high_low_two_models.tex
\begin{table*}[!ht]
    \centering
    \setlength\tabcolsep{4pt} 
    \resizebox{\textwidth}{!}{
    
    \begin{tabular}{llcccccccccccccc}
    \toprule
        \multirow{2}{*}{\textbf{Model}} & \multirow{2}{*}{\textbf{Prompt type}} & \multicolumn{2}{c}{\textbf{MGSM}} & \multicolumn{2}{c}{\textbf{XCOPA}} & \multicolumn{2}{c}{\textbf{XNLI}} & \multicolumn{2}{c}{\textbf{PAWS-X}} & \multicolumn{2}{c}{\textbf{MKQA}} & \multicolumn{2}{c}{\textbf{XL-Sum}} & \multicolumn{2}{c}{\textbf{AVG}} \\
        & &  high & low & high & low & high & low & high & low & high & low & high & low & high & low  \\
        \midrule
        \multirow{7}{*}{ChatGPT} 
        &\nativebasic & 44.4 & 19.4 & 84.6 & 69.7 & 56.9 & 48.6 & 51.6 & 40.6 & 35.1 & 36.4 & 32.5 & 29.9 & 50.8 & 40.8 \\ 
        &\enbasic & 50.3 & 27.3 & 88.3 & 73.3 & 64.6 & 61.8 & 64.3 & 50.4 & 37.4 & 33.3 & \textbf{33.3} & \textbf{30.0} & 56.4 & 46.0 \\ 
       & \nativecot & 65.1 & 27.1 & 84.1 & 69.8 & 54.9 & 47.4 & 51.6 & 43.4 & 35.5 & 35.1 & 31.9 & 27.9 & 53.8 & 41.8 \\ 
        &\encot & 70.5 & 47.1 & 89.9 & 75.9 & 60.2 & 53.6 & 63.7 & 51.2 & \textbf{43.3} & 41.2 & 30.0 & 28.6 & 59.6 & 49.6 \\ 
        &\xlt & 70.4 & 50.1 & 89.3 & 76.8 & 60.6 & 58.1 & 59.7 & 58.2 & 37.7 & 37.5 & 22.8 & 26.1 & 56.7 & 51.1 \\ 
        &\transg & \textbf{74.7} & \textbf{72.7} &\textbf{90.3} & \textbf{83.2} & \textbf{62.4} & \textbf{59.1} & 68.2 & 62.0 & 42.5 & \textbf{48.3} & 30.6 & 28.9 & \textbf{61.4} & \textbf{59.0}\\ 
        &\transn & 65.6 & 54.1 & 85.7 & 78.2 & 60.5 & 58.2 &\textbf{68.4} & \textbf{63.4} & 35.4 & 43.6 & 28.4 & 27.7 & 57.3 & 54.2 \\ 
         \midrule
        \multirow{7}{*}{Llama-2-70B-Chat} 
        &\nativebasic & 35.7 & 5.6 & 64.2 & 48.0 & 43.0 & 36.0 & 53.3 & 50.4 & 28.9 & 10.4 & 30.1 & 26.8 & 42.5 & 29.5 \\ 
        &\enbasic & 42.5 & 7.7 & 70.7 & 52.0 & 52.7 & 41.9 & 61.9 & 52.8 & 25.7 & 21.5 & 30.2 & 35.3 & 47.3 & 35.2 \\ 
        &\nativecot & 35.5 & 5.6 & 65.3 & 46.8 & 41.0 & 35.6 & 56.0 & 49.6 & 25.3 & 9.9 & 26.0 & 25.2 & 41.5 & 28.8 \\ 
        &\encot & 45.6 & 7.0 & 80.7 & 56.3 & 52.7 & 40.9 & 66.5 & 57.0 & 32.7 & 25.7 & 29.8 & 32.0 & 51.3 & 36.5 \\ 
        &\xlt & 49.0 & 8.4 & 76.4 & 54.7 & \textbf{57.3} & 48.4 & 56.6 & 51.6 & 26.5 & 26.7 & 19.3 & 11.5 & 47.5 & 33.6 \\ 
       & \transg & \textbf{55.5} & \textbf{50.0} & \textbf{86.3} & \textbf{79.7} & 55.3 & \textbf{53.0} & 69.4 & \textbf{64.2} & \textbf{38.7} & \textbf{43.1} & \textbf{33.1} & \textbf{36.7} & \textbf{56.4} & \textbf{54.4}\\ 
        &\transn & 46.5 & 39.7 & 83.3 & 75.6 & 53.7 & 51.0 & \textbf{70.5} & 62.4 & 17.8 & 24.7 & 32.4 & 36.2 & 50.7 & 48.3 \\ 
    \bottomrule
    \end{tabular}
    }
    \caption{Average scores of the high-resource languages and low-resource languages for the six benchmarks in zero-shot setting. 
    The best result for each model is in \textbf{bold}.
    }
    \label{tab:main_NLP_HL_two_models}
\end{table*}

%% file: latex/tables/analysis_distance.tex
\begin{table}[!t]
    \centering
    \setlength\tabcolsep{4pt} 
    \resizebox{\linewidth}{!}{
    \begin{tabular}{llllll}
        \toprule
        \textbf{Prompt type} & \textbf{SYN} & \textbf{GEO} & \textbf{INV} & \textbf{GEN} & \textbf{PHON} \\
        \midrule
        \underline{\texttt{ChatGPT}} \\
        \nativebasic & -0.786* & -0.336 & 0.323 & -0.403 & -0.044 \\ 
        \enbasic & -0.820* & -0.160 & 0.527 & -0.299 & 0.020 \\ 
        \nativecot & -0.795* & -0.184 & 0.479 & -0.313 & 0.045 \\ 
        \encot & -0.841* & -0.286 & 0.339 & -0.436 & -0.034 \\ 
        \xlt & -0.787* & -0.113 & 0.445 & -0.284 & 0.117 \\ 
        \midrule
        \multicolumn{2}{l}{\underline{\texttt{Llama-2-70B-Chat}}} \\
        \nativebasic & -0.688* & -0.369 & 0.250 & -0.323 & -0.044 \\ 
        \enbasic & -0.782* & -0.512 & 0.134 & -0.513 & -0.226 \\ 
        \nativecot & -0.706* & -0.403 & 0.231 & -0.475 & -0.105 \\ 
        \encot & -0.737* & -0.510 & 0.206 & -0.445 & -0.219 \\ 
        \xlt & -0.697* & -0.432 & 0.266 & -0.423 & -0.153 \\ 
        \bottomrule
    \end{tabular}
    }
    \caption{Pearson correlation coefficient between MGSM accuracy and five language distances between English and that language. A lower value indicates higher correlation due to the negative coefficients.(*p < 0.05, two-tailed)}
    \label{tab:distance}
\end{table}

%% file: latex/tables/result_bilingual_NLP.tex
\begin{table*}[!t]
    \centering
    \small
    \setlength\tabcolsep{2pt}
    \resizebox{\linewidth}{!}{%
    \begin{tabular}{lccccccc c ccccccc}
    \toprule
    \multirow{2}{*}{Prompt type} 
      & \multicolumn{7}{c}{\texttt{Qwen1.5-72B-Chat}} 
      & \multicolumn{1}{c}{} 
      & \multicolumn{7}{c}{\texttt{Yi-34B-Chat}} \\
    \cmidrule(lr){2-8} \cmidrule(lr){10-16}
      & MGSM & XCOPA & XNLI & PAWS-X & MKQA & XL-Sum & AVG 
      & 
      & MGSM & XCOPA & XNLI & PAWS-X & MKQA & XL-Sum & AVG \\
    \midrule
    \nativebasic 
      & 78.8 & 93.0 & 55.8 & 71.8 & 36.6 & \textbf{41.3} & 62.9 
      & 
      & 63.2 & 92.6 & 46.0 & 43.6 & 13.4 & \textbf{36.9} & 49.3 \\
    \enbasic 
      & 77.2 & 97.0 & 73.0 & \textbf{73.0} & 32.7 & 39.7 & 65.4 
      & 
      & 66.8 & 93.6 & 52.6 & \textbf{74.6} & 15.5 & 35.1 & 56.4 \\
    \nativecot 
      & \textbf{83.2} & 95.8 & 46.4 & 72.2 & 35.8 & 39.5 & 62.1 
      & 
      & 65.2 & 91.8 & 42.6 & 43.6 & 13.0 & 36.6 & 48.8 \\
    \encot 
      & 81.6 & 97.2 & 71.2 & 70.6 & 34.9 & 38.6 & \textbf{65.7} 
      & 
      & 70.0 & 93.6 & 48.2 & 74.8 & 12.1 & 33.1 & 55.3 \\
    \xlt 
      & 78.4 & \textbf{97.8} & \textbf{77.4} & 67.6 & 20.8 & 35.3 & 62.9 
      & 
      & 56.0 & 93.2 & \textbf{69.2} & 65.6 & 7.5 & 31.3 & 53.8 \\
    \transg 
      & 81.6 & 94.6 & 63.8 & 68.4 & \textbf{45.7} & 31.3 & 64.2 
      & 
      & \textbf{71.2} & \textbf{94.0} & 49.6 & 70.8 & \textbf{24.5} & 36.3 & \textbf{57.7} \\
    \transn 
      & 58.8 & 88.2 & 61.4 & 70.4 & 32.0 & 28.5 & 56.5 
      & 
      & 56.0 & 86.6 & 48.8 & 68.2 & 22.9 & 28.5 & 51.8 \\
    \bottomrule
    \end{tabular}%
    }
    \caption{Scores of the two non-English-centric LLMs on NLP tasks for the Chinese language. The best result for each model is in \textbf{bold}.}
    \label{tab:bilingual_NLP}
\end{table*}

%% file: latex/sections/related_work.tex
\section{Related Work}
\paragraph{Multilingual Evaluation.}
Since the release of ChatGPT, the evaluation of LLMs has attracted the attention of the research community\cite{qin_is_2023, bang_multitask_2023}. \citealp{shi_language_2022} evaluated LLMs on MGSM and found that the models demonstrated strong multilingual reasoning
 capabilities, even for low-resource languages. \citealp{bang_multitask_2023} evaluated ChatGPT on 23 datasets covering 8 NLP tasks. They found that ChatGPT failed to generalize its capabilities to non-Latin scripts. To cover tasks, \citealp{ahuja_mega_2023} evaluated ChatGPT and GPT-4 on 16 NLP datasets across 70 languages and compared them with state-of-the-art non-autoregressive models. Concurrently, \citealp{lai_chatgpt_2023} evaluated ChatGPT on 7 different tasks across 37 diverse languages. However, these evaluations are primarily limited to standard NLP tasks and largely overlook real-world scenarios and cultural knowledge \cite{fung_massively_2024}, which are crucial for understanding the practical applicability of LLMs.

\paragraph{Multilingual Prompting Strategies.} 
The translate-test is a popular technique used to refine the performance of multilingual NLP benchmarks \cite{conneau_xnli_2018,ponti_xcopa_2020,artetxe_revisiting_2023,moghe_extrinsic_2023,qi_enhancing_2022,huang_zero-shot_2022}. In the era of LLMs, various strategies have been developed to enhance the performance of LLMs using multilingual datasets. \citealp{shi_language_2022} discovered that 
\encot outperforms \nativecot. \citealp{huang_not_2023} introduced cross-lingual-thought prompting to minimize language disparities. In parallel, \citealp{qin_cross-lingual_2023} introduced cross-lingual prompting, and \citealp{etxaniz_multilingual_2023} suggested self-translate to elevate their performances. Effective in translating prompts into English, these methods excel in NLP tasks but remain uncertain in real-world applications. Their success hinges on the English-centric nature of the LLMs. Our study evaluates translation effectiveness across NLP tasks, real user queries, and non-English-centric LLMs, revealing the limitations of these methods.

%% file: latex/sections/appendix.tex
\section{Appendix}
\label{sec:appendix}

\subsection{Translation for NLP Tasks}\label{Appendix:NLP}
This section presents more details about the setups and results for the experiments on NLP tasks.

\subsubsection{Details about NLP Benchmarks}
Here are the detailed descriptions of the NLP benchmarks: 
\paragraph{Arithmetic Reasoning}
The MGSM \cite{shi_language_2022} benchmark includes mathematical problems from grade school and requires the model to compute the accurate solution. It spans 10 languages, and we use the accuracy score for assessment.

\paragraph{Commonsense Reasoning}
The XCOPA benchmark \cite{ponti_xcopa_2020} consists of a single premise and two choices. The goal is to identify which choice is the cause or effect of the premise. It covers 11 languages from various families, with an accuracy score used for evaluation.

\paragraph{Natural Language Inference}
The XNLI \cite{conneau_xnli_2018} benchmark includes one premise and one hypothesis. The model's job is to determine if the hypothesis is entailed, contradicted, or neutral based on the premise. It covers 15 languages, and we evaluate it using the accuracy score.

\paragraph{Paraphrase Identification}
The PAWS-X~\citep{yang_paws-x_2019} benchmark consists of two sentences and requires the model to judge whether they are paraphrases. It covers 7 languages, and we assess based on accuracy score.

\paragraph{Question Answering}
The MKQA dataset \citep{longpre_mkqa_2021} contains open-domain questions that require predicting short answers. Questions that are unanswerable or excessively long to have a specific answer are not considered during evaluation. This dataset covers 25 languages, with our focus on 9 languages: de, es, fr, ja, ru, th, tr, vi, and zh. We assess the model's performance using the token overlap F1 score.

\paragraph{Summarization}
The XL-Sum \cite{hasan_xl-sum_2021} benchmark requires the model to condense a lengthy news article into a brief summary. It covers 44 languages, and we select a subset of 5 languages: es, fr, tr, vi, and zh. 
We use the ROUGE-1 score for evaluation.

\subsubsection{More LLMs for Experiment}
Besides ChatGPT and Llama-2-70B-Chat, we have also evaluated the NLP tasks with the following models: 
\begin{itemize}
    \item Mistral-7B-Instruct (v0.2). 
    This model is the instructed version of Mistral-7B~\cite{jiang_mistral_2023}. 
    \item Llama-2-13B-chat, which is a chat model in Llama-2 family \cite{touvron_llama_2023}. 
    \item bloomz-7b1, which is a model fine-tuned with multiple tasks, including some multilingual tasks \cite{muennighoff_crosslingual_2023}.
\end{itemize}

\subsubsection{More Details about Prompt Strategies} \label{app:nlp_prompt}
An example of various prompting strategies is shown in Table \ref{tab:prompt_example}.
The prompts of \enbasic for each task are shown in Table \ref{tab:basic-prompt}, which are adapted from \citealp{huang_not_2023}. The translation template for self-translate with LLMs is: \\\textit{Translate the following question from \{language\} to English: \\\{question\}\\
Don't answer the question, just translate it!} \\
The prompt templates for other prompting strategies and the instructions for output formats are designed according to the descriptions in Section \ref{sec:strategy}.

\subsubsection{Additional Results}\label{appendix:results}
The average performances for high-resource and low-resource languages are shown in Table \ref{tab:main_NLP_HL}.
Table \ref{tab:result_mgsm}, Table \ref{tab:result_xcopa}, Table \ref{tab:result_XNLI}, Table \ref{tab:result_paws}, Table \ref{tab:result_mkqa} and Table \ref{tab:result_xlsum} shows the detailed results for MGSM, XCOPA, XNLI, PAWS-X, MKQA and XL-Sum, respectively. In addition to the finding in Section \ref{NLP_results}, We find \xlt exhibits competitive performance in reasoning tasks; however, its performance in generation tasks is less impressive. Our findings indicate that when employing the \xlt prompting strategy, ChatGPT declined to answer 26.4\% of the questions in the XL-Sum tasks, responding with ``\textit{I'm sorry, I cannot \ldots}'' This refusal pattern was not observed when utilizing other prompting strategies. For open-source models, while we did not observe a refusal pattern, they do not follow the instructions properly, which also degrades their performance with \xlt.

\input{latex/tables/zero-shot-prompts}

\input{latex/tables/templates}
\input{latex/tables/main_results_high_low}

\input{latex/tables/result_mgsm}

\input{latex/tables/result_xcopa}
\input{latex/tables/results_xnli}

\input{latex/tables/result_paws-x}

\input{latex/tables/result_mkqa}

\input{latex/tables/result_xlsum}

\subsection{Translation for Real User Queries}
\label{app:shareGPT}
\begin{figure}
    \centering
    \includegraphics[width=0.95\linewidth]{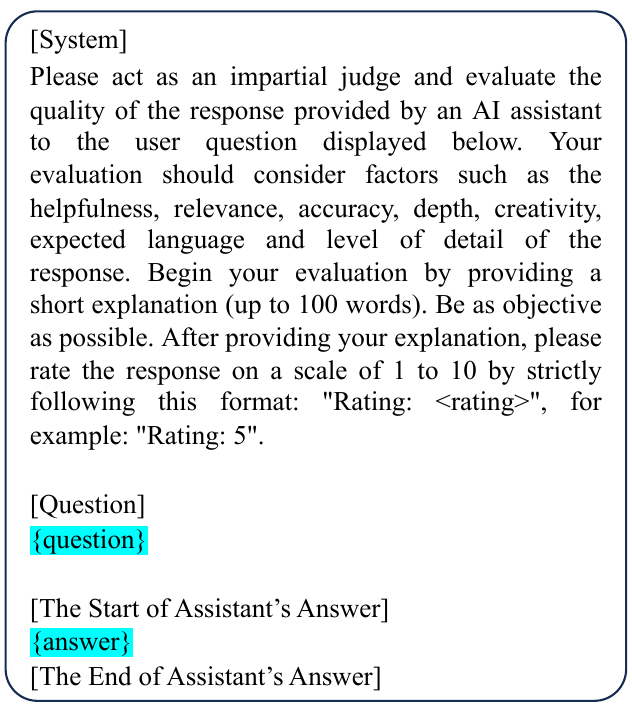}
    \caption{The LLM-as-a-judge prompt for GPT-4o.}
    \label{fig:judge}
\end{figure}

\begin{figure*}
    \centering
    \includegraphics[width=0.95\linewidth]{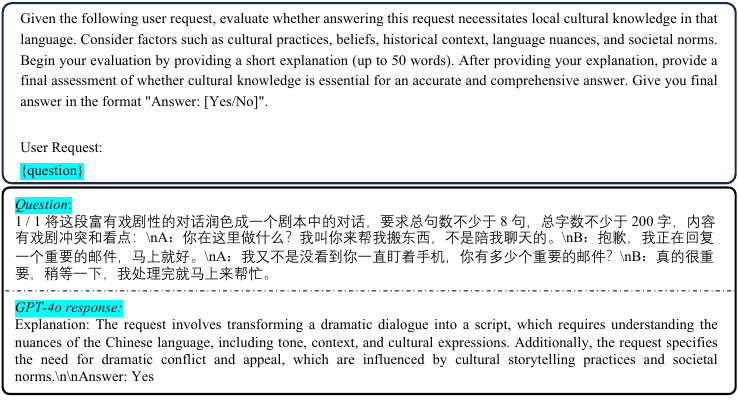}
    \caption{Prompt template to check whether answering a request needs local cultural knowledge (upper) and one Chinese example (lower).}
    \label{fig:prompt_culture}
\end{figure*}

The prompt used to assess the response quality is shown in Figure \ref{fig:judge}. When GPT-4o is prompted with this, it assigns a score ranging from 1 to 10 to each response.
Figure \ref{fig:prompt_culture} illustrates the prompt used to determine if responding to a request requires local cultural knowledge. The Chinese case shows that GPT-4o can identify if queries require knowledge of local culture with explanations and the final answer. 

\begin{figure*}[t]
    \centering

    \begin{subfigure}[b]{0.48\textwidth}
        \includegraphics[width=\textwidth]{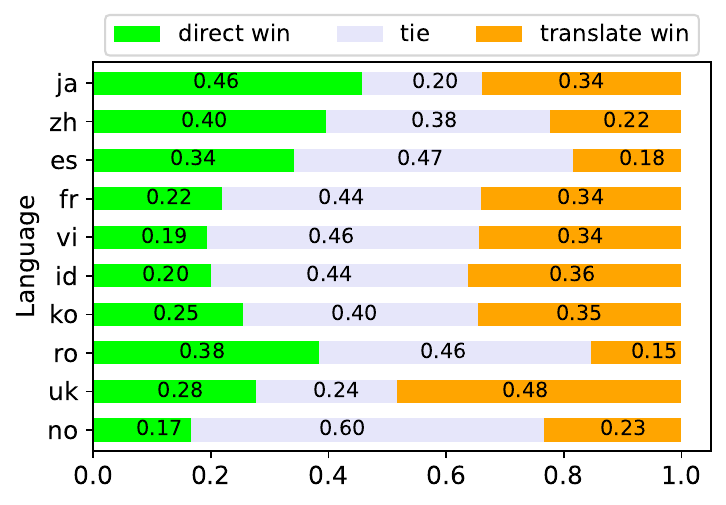}
        \caption{Win rate with ChatGPT (w/ cultural knowledge)}
        \label{fig:sub1}
    \end{subfigure}
    \hfill 
    \begin{subfigure}[b]{0.48\textwidth}
        \includegraphics[width=\textwidth]{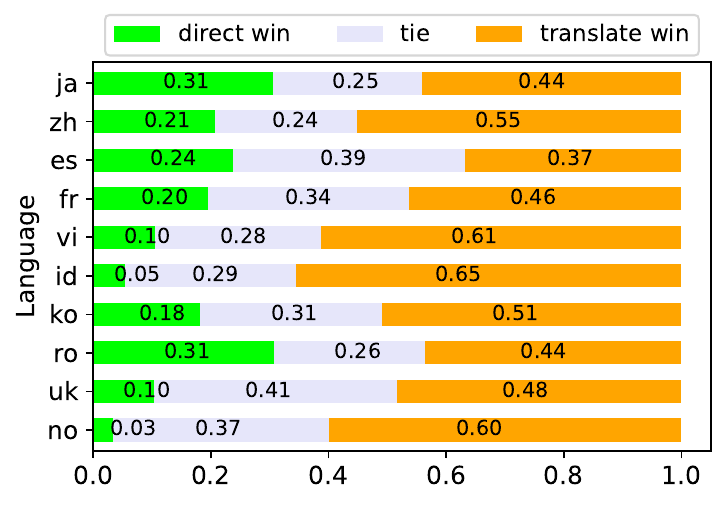}
        \caption{Win rate with Llama-2-70B-Chat (w/ cultural knowledge)}
        \label{fig:sub2}
    \end{subfigure}
    \caption{Win rate comparison for each language using ChatGPT and Llama-2-70B-Chat for the subsets of shareGPT with  cultural knowledge.}
    \label{fig:results_shareGPT_subset}
\end{figure*}

We also analyzed the performance of shareGPT subsets with cultural knowledge only. As shown in Figure \ref{fig:results_shareGPT_subset}, the behaviors across languages and models are inconsistent. ChatGPT shows different behaviors for high-resource and low-resource languages. For high-resource languages like Japanese, Chinese, and Spanish, prompting with original queries has a higher win rate. For low-resource languages, translation is often a better option. In contrast, Llama-2-70B-Chat shows a higher win rate for all languages.

\subsubsection{Additional Results} \label{sec:real_query_more_judges}
\begin{figure*}[t]
    \centering
    \begin{subfigure}[b]{0.48\textwidth}
        \includegraphics[width=\textwidth]{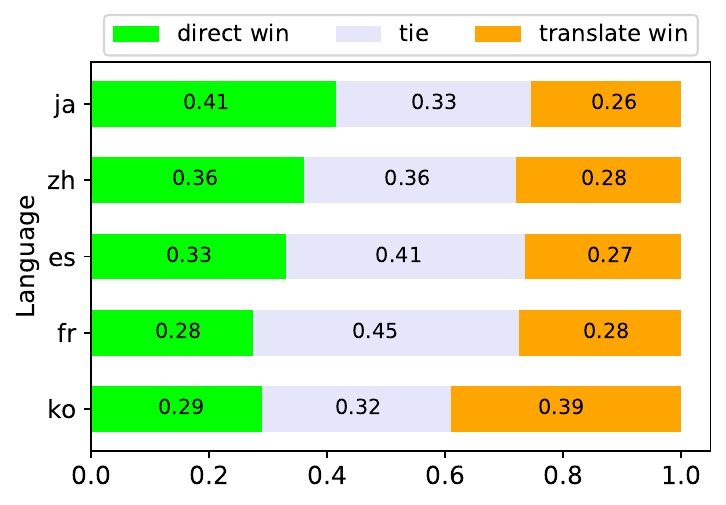}
        \caption{ChatGPT judged by GPT-4o}
    \end{subfigure}
    \begin{subfigure}[b]{0.48\textwidth}
        \includegraphics[width=\textwidth]{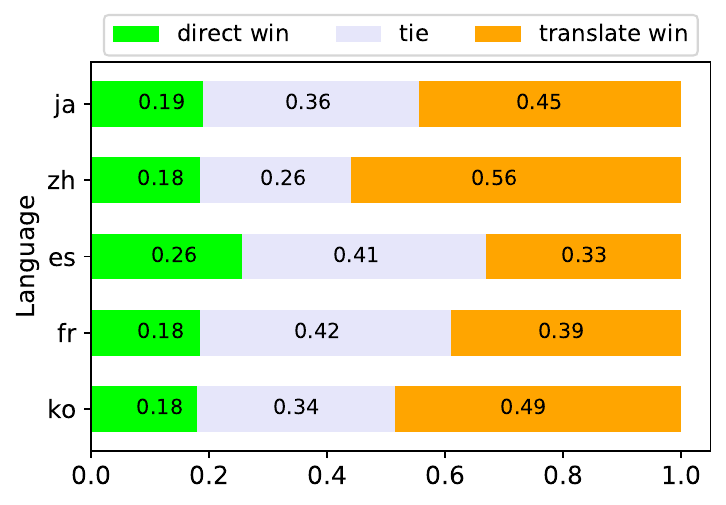}
        \caption{Llama-2-70B-Chat judged by GPT-4o}
    \end{subfigure}
    \begin{subfigure}[b]{0.48\textwidth}
        \includegraphics[width=\textwidth]{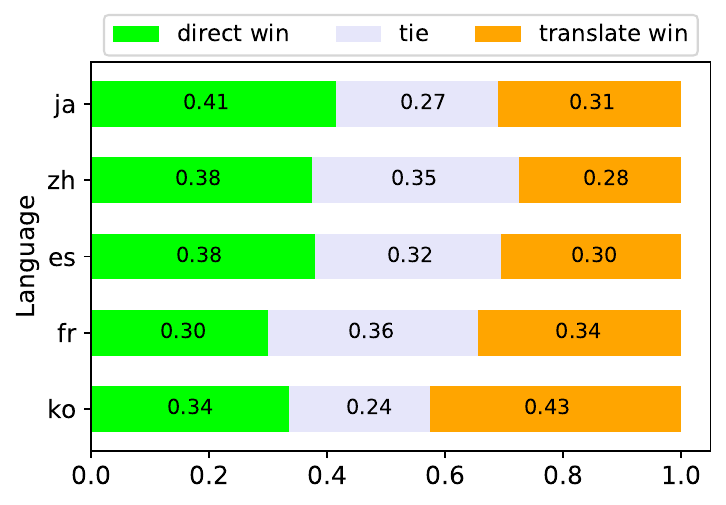}
        \caption{ChatGPT judged by Claude-3.5-Sonnet}
    \end{subfigure}
        \begin{subfigure}[b]{0.48\textwidth}
        \includegraphics[width=\textwidth]{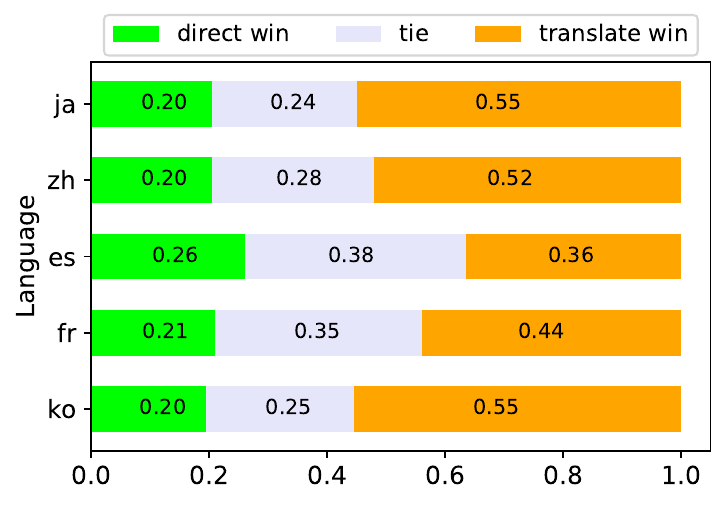}
        \caption{Llama-2-70B-Chat judged by Claude-3.5-Sonnet}
    \end{subfigure}
    \begin{subfigure}[b]{0.48\textwidth}
        \includegraphics[width=\textwidth]{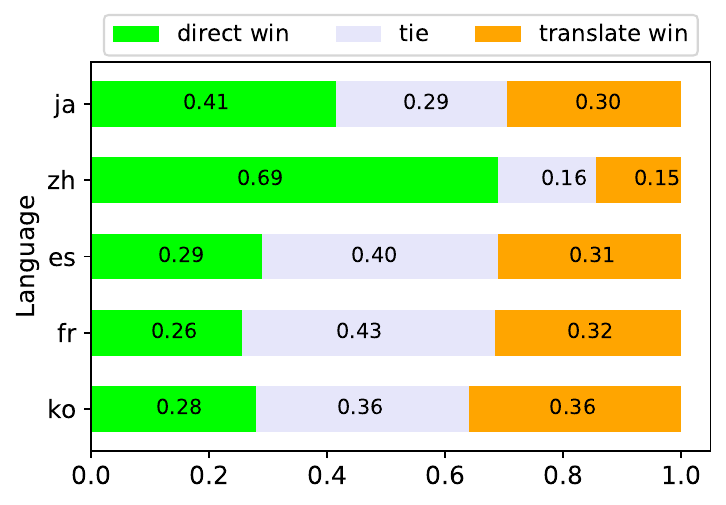}
        \caption{ChatGPT judged by Gemini-1.5-Pro}
    \end{subfigure}
    \begin{subfigure}[b]{0.48\textwidth}
        \includegraphics[width=\textwidth]{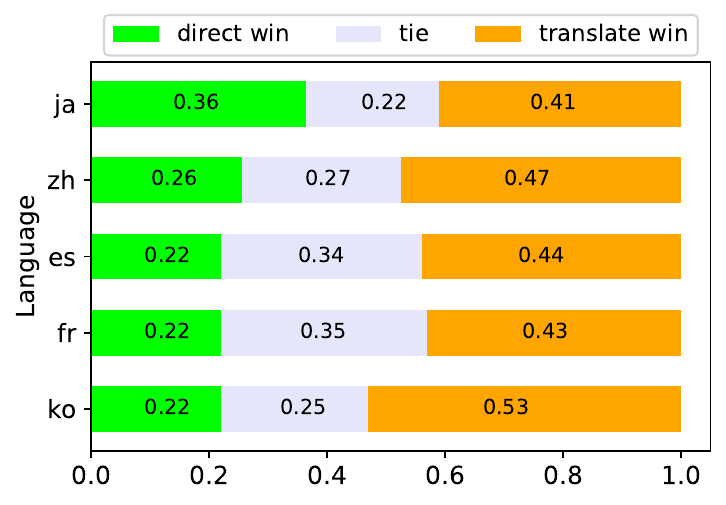}
        \caption{Llama-2-70B-Chat judged by Gemini-1.5-Pro}
    \end{subfigure}
    \caption{Win rate comparison for five languages using ChatGPT and Llama-2-70B-Chat judged with three advanced LLMs.}
    \label{fig:results_shareGPT_more_judges}
\end{figure*}

In Section \ref{sec:real_query_setup}, we randomly select 100 requests for each language and evaluate the quality of the responses generated by GPT-4o. To ensure a more rigorous and comprehensive analysis, we conduct additional experiments under the following conditions: we heuristically filter queries using GPT-4o to ensure their validity, select 200 queries per language from the filtered set, and employ multiple judge models. Due to an insufficient number of available queries in other languages, we limit our evaluation to Japanese (ja), Chinese (zh), Spanish (es), French (fr), and Korean (ko). For the judging process, we use not only GPT-4o but also Claude-3.5-Sonnet and Gemini-Pro-1.5 to provide a more diverse assessment. The results are presented in Figure \ref{fig:results_shareGPT_more_judges}. ChatGPT performs better when given direct prompts in languages such as Japanese and Chinese, whereas Llama-2-70B-Chat consistently achieves higher performance with translated prompts. These findings align with those discussed in Section \ref{sec:real_query_results}.

\begin{table*}[t]
    \centering
    \small
    \begin{tabular}{lllllllllll}
        \toprule
        \textbf{Language} & \textbf{ja} & \textbf{zh} & \textbf{es} & \textbf{fr} & \textbf{vi} & \textbf{id} & \textbf{ko} & \textbf{ro} & \textbf{uk} & \textbf{no} \\
        \midrule
         \textbf{Ratio (\%)} & 59 & 58 & 38 & 41 & 67 & 55 & 55 & 74 & 30 & 57 \\
        \bottomrule
    \end{tabular}
    \caption{The percentage of the questions that necessitate local cultural knowledge.}
    \label{tab:ratio_culture}
\end{table*}

\subsection{Layerwise Language Distribution in Larger Model}\label{sec:appen_layer}

\begin{figure*}[t]
    \centering
    \begin{subfigure}[b]{0.49\textwidth}
        \includegraphics[width=\textwidth]{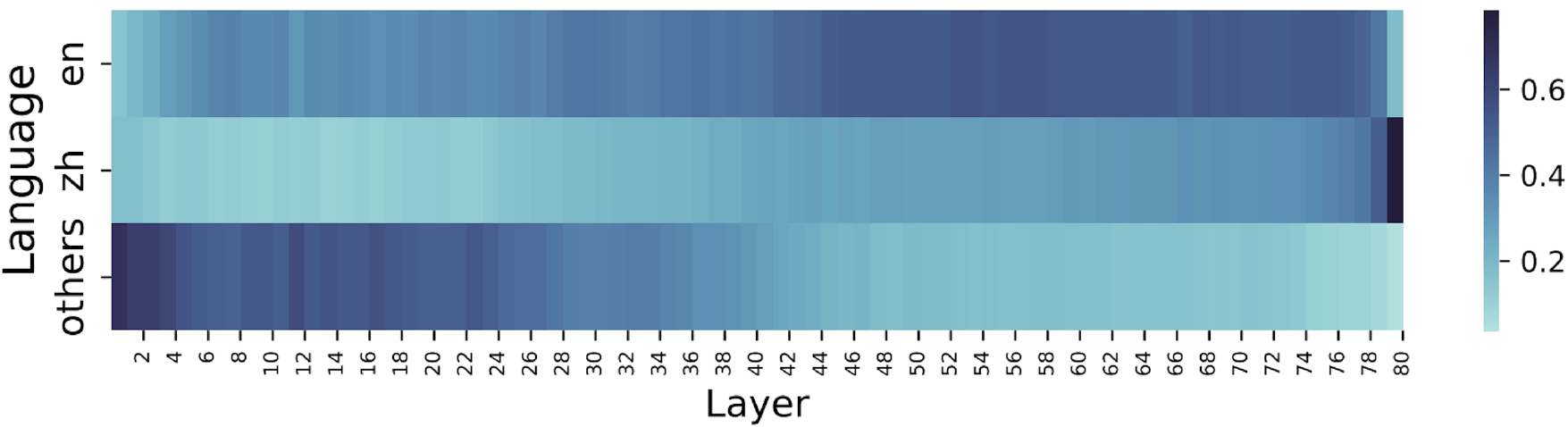}
        \caption{Llama-2-70B-Chat}
        \label{fig:layer1}
    \end{subfigure}
    \hfill 
    \begin{subfigure}[b]{0.49\textwidth}
        \includegraphics[width=\textwidth]{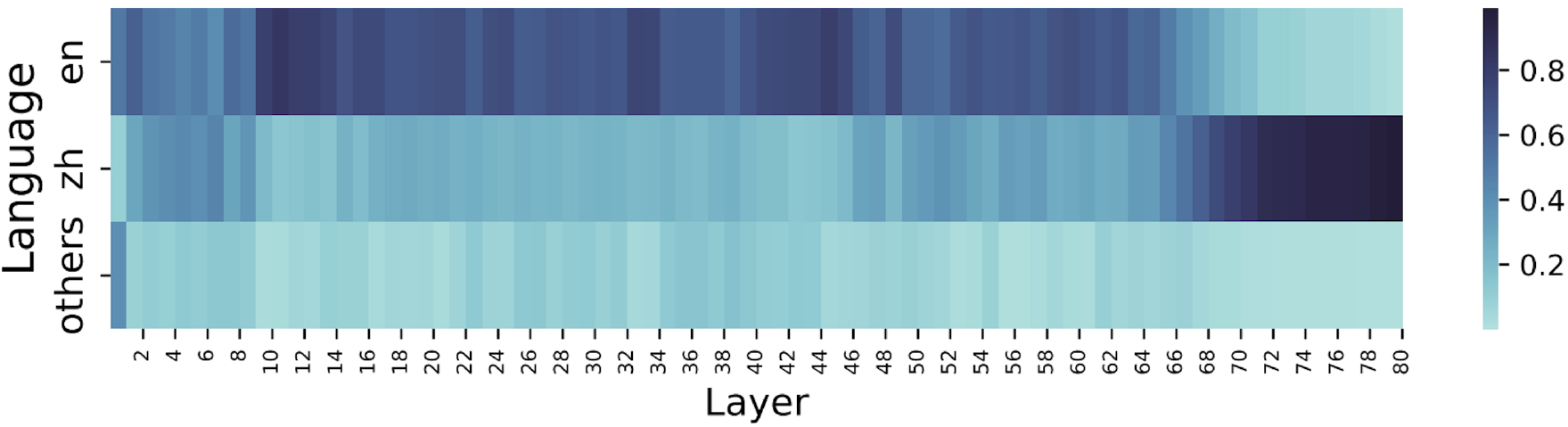}
        \caption{Qwen1.5-72B-Chat}
        \label{fig:layer2}
    \end{subfigure}
    \caption{Layerwise language distribution for (a) Llama-2-70b-Chat and (b) Qwen1.5-72B-Chat with Chinese prompts.}
    \label{fig:layerwise_lang_2}
\end{figure*}

Figure \ref{fig:layerwise_lang_2} illustrates the layerwise language distribution in larger models, including Llama-2-70B-Chat and Qwen1.5-72B-Chat. Llama-2-70B-Chat exhibits the same phenomenon as its smaller counterpart, Llama-2-7B-chat, with diverse languages represented in its hidden states. In contrast to Qwen1.5-7B-Chat, the hidden representations of Qwen1.5-72B-Chat incorporate both Chinese and English until the last few layers, possibly reflecting the challenges of constructing such a large model using Chinese exclusively for hidden representations. Nevertheless, it still represents its hidden states more in Chinese than Llama-2-70B-Chat.

%% file: latex/tables/zero-shot-prompts.tex
\begin{table*}[t]
    \centering \small
    \begin{tabular}{p{0.19\textwidth}p{0.68\textwidth}}
    \toprule
    \bf Original Question & \begin{CJK*}{UTF8}{gbsn}制作一件袍子需要 2 匹蓝色纤维布料和这个数量一半的白色纤维布料。它一共需要用掉多少匹布料\end{CJK*} \\
    \midrule
    \nativebasic & \{\textbf{Original Question} \} \newline
    \chinese{您的最终答案的格式应为："答案: <阿拉伯数字>".} \\
    \midrule
    \enbasic & \{\textbf{Original Question} \} \newline
    You should format your final answer as "Answer: <Arabic numeral>".\\ 
    \midrule
    \nativecot & \{\textbf{Original Question} \} \newline
    \chinese{让我们一步步思考。} \newline
    \chinese{您的最终答案的格式应为："答案: <阿拉伯数字>".} \\
    \midrule
    \encot & \{\textbf{Original Question} \}\newline
    Let's think step by step in English. \newline 
    You should format your final answer as "Answer: <Arabic numeral>".\\ 
    \midrule
    \xlt & I want you to act as an arithmetic reasoning expert for Chinese. \newline Request: \{\textbf{Original Question}\}\newline You should retell the request in English. \newline You should do step-by-step answer to obtain a number answer.\newline You should step-by-step answer the request. \newline You should tell me the answer in this format 'Answer :'. \\
    \midrule    
    \transg & Crafting a robe requires 2 bolts of blue fiber cloth and half that amount of white fiber cloth. How many pieces of fabric will it use in total?\newline Let's think step by step.\newline You should format your final answer as "Answer: <Arabic numeral>". \\
    \midrule
    \transn & To make a robe, two pieces of blue fiber and half of that amount of white fiber are needed. How many pieces of fabric does it take to make?\newline Let's think step by step.\newline You should format your final answer as "Answer: <Arabic numeral>". \\
    \bottomrule 
    \end{tabular}
    \caption{An example of zero-shot prompts for a Chinese problem. For \nativebasic, \enbasic, \nativecot, \encot and \xlt, we provide the original Chinese question as input and expect an answer in the corresponding format; for \transg and \transn, we input the translated question in English, and expect a step-by-step solution in English. 
    To obtain the desirable output format, we instruct the models to output in specific format. 
}
    \label{tab:prompt_example}
\end{table*}

%% file: latex/tables/templates.tex
\begin{table*}[t]
\centering
\small
\begin{tabular}{lrp{0.73\textwidth}}
\toprule
\textbf{Benchmark} & \textbf{\#Test} & \textbf{Basic Prompt} \\ 
\midrule
MGSM & 250 & \texttt{\{problem\}} \\
\midrule[0.3pt]
XCOPA & 500 & Here is a premise: \texttt{\{premise\}}. What is the \texttt{\{question\}}? Help me pick the more plausible option: -choice1: \texttt{\{choice1\}}, -choice2: \texttt{\{choice2\}} \\
\midrule[0.3pt]
XNLI & 500 & \texttt{\{premise\}} Based on previous passage, is it true that \texttt{\{hypothesis\}}? 1: Yes, 2: No, or 3: Maybe? \\ 
\midrule[0.3pt]
PAWS-X & 500 & Sentence 1: \texttt{\{sentence1\}} Sentence 2: \texttt{\{sentence2\}} Question: Does Sentence 1 paraphrase Sentence 2? 1: Yes, 2: No? \\
\midrule[0.3pt]
MKQA & 500 & Answer the question in one or a few words in \texttt{\{target\_language\}}: \texttt{\{question\}}? \\
\midrule[0.3pt]
XL-Sum & 500 & \texttt{\{article\}} Summarize the article. \\
\bottomrule
\end{tabular}
\caption{Template of \enbasic for each benchmark. \#Test denotes the number of samples in the test set.}
\label{tab:basic-prompt}
\end{table*}

%% file: latex/tables/main_results_high_low.tex
\begin{table*}[!ht]
    \centering
    \setlength\tabcolsep{4pt} 
    \resizebox{\textwidth}{!}{
    
    \begin{tabular}{llcccccccccccccc}
    \toprule
        \multirow{2}{*}{\textbf{Model}} & \multirow{2}{*}{\textbf{Prompt type}} & \multicolumn{2}{c}{\textbf{MGSM}} & \multicolumn{2}{c}{\textbf{XCOPA}} & \multicolumn{2}{c}{\textbf{XNLI}} & \multicolumn{2}{c}{\textbf{PAWS-X}} & \multicolumn{2}{c}{\textbf{MKQA}} & \multicolumn{2}{c}{\textbf{XL-Sum}} & \multicolumn{2}{c}{\textbf{AVG}} \\
        & &  high & low & high & low & high & low & high & low & high & low & high & low & high & low  \\
        \midrule
        \multirow{7}{*}{ChatGPT} 
        &\nativebasic & 44.4 & 19.4 & 84.6 & 69.7 & 56.9 & 48.6 & 51.6 & 40.6 & 35.1 & 36.4 & 32.5 & 29.9 & 50.8 & 40.8 \\ 
        &\enbasic & 50.3 & 27.3 & 88.3 & 73.3 & 64.6 & 61.8 & 64.3 & 50.4 & 37.4 & 33.3 & \textbf{33.3} & \textbf{30.0} & 56.4 & 46.0 \\ 
       & \nativecot & 65.1 & 27.1 & 84.1 & 69.8 & 54.9 & 47.4 & 51.6 & 43.4 & 35.5 & 35.1 & 31.9 & 27.9 & 53.8 & 41.8 \\ 
        &\encot & 70.5 & 47.1 & 89.9 & 75.9 & 60.2 & 53.6 & 63.7 & 51.2 & \textbf{43.3} & 41.2 & 30.0 & 28.6 & 59.6 & 49.6 \\ 
        &\xlt & 70.4 & 50.1 & 89.3 & 76.8 & 60.6 & 58.1 & 59.7 & 58.2 & 37.7 & 37.5 & 22.8 & 26.1 & 56.7 & 51.1 \\ 
        &\transg & \textbf{74.7} & \textbf{72.7} &\textbf{90.3} & \textbf{83.2} & \textbf{62.4} & \textbf{59.1} & 68.2 & 62.0 & 42.5 & \textbf{48.3} & 30.6 & 28.9 & \textbf{61.4} & \textbf{59.0}\\ 
        &\transn & 65.6 & 54.1 & 85.7 & 78.2 & 60.5 & 58.2 &\textbf{68.4} & \textbf{63.4} & 35.4 & 43.6 & 28.4 & 27.7 & 57.3 & 54.2 \\ 
         \midrule
        \multirow{7}{*}{bloomz-7b1} 
         &\nativebasic & 1.6 & 0.9 & 36.5 & 18.9 & 3.7 & 11.8 & - & - & 7.1 & 10.5 & - & - & 12.2 & 10.5 \\ 
       & \enbasic & 1.9 & 2.2 & 67.5 & 55.2 & \textbf{48.2} & 40.7 & - & - & 11.8 & 6.5 & - & - & 32.4 & 26.2 \\ 
        &\nativecot & 1.0 & 1.4 & 37.9 & 17.3 & 1.2 & 13.5 & - & - & 5.2 & 11.1 & - & - & 11.3 & 10.8 \\ 
        &\encot & 1.7 & 1.6 & 61.3 & 52.8 & 37.6 & 34.7 & - & - & 10.0 & 6.9 & - & - & 27.7 & 24.0 \\ 
       & \xlt & 1.9 & 1.5 & 58.6 & 49.2 & 35.4 & 35.3 & - & - & 8.6 & 5.9 & - & - & 26.1 & 23.0 \\ 
       & \transg & \textbf{2.5} & \textbf{3.0} & \textbf{67.5} & \textbf{62.8} & 44.4 &\textbf{44.2} & - & - & \textbf{15.6} & \textbf{23.0} & - & - & \textbf{32.5} & \textbf{33.2} \\ 
        &\transn & 2.0 & 2.9 & 64.3 & 61.2 & 44.1 & 43.6 & - & - & 12.8 & 21.3 & - & - & 30.8 & 32.2 \\ 
        \midrule
        \multirow{7}{*}{Mistral-7B-Instruct} 
        &\nativebasic & 15.5 & 4.9 & 69.7 & 50.0 & 50.6 & 37.0 & 44.6 & 44.8 & 7.8 & 8.1 & 26.3 & 24.4 & 35.7 & 28.2 \\ 
        &\enbasic & 33.7 & 8.8 & 42.5 & 33.8 & 55.5 & 46.2 & 47.0 & 46.6 & 6.8 & 8.0 & 21.7 & 21.1 & 34.5 & 27.4 \\ 
        &\nativecot & 23.1 & 8.0 & 67.7 & 49.9 & 50.2 & 38.3 & 44.3 & 44.2 & 7.7 & 8.2 & 25.5 & 21.1 & 36.4 & 28.3 \\ 
        &\encot & 37.3 & 13.1 & 50.9 & 38.9 & 54.2 & 46.8 & 46.6 & 46.4 & 11.3 & 12.0 & 18.7 & 18.8 & 36.5 & 29.3 \\ 
        &\xlt & \textbf{43.0} & 15.0 & \textbf{78.3} & \textbf{57.9} & 48.4 & 44.3 & 47.9 & 47.2 & 9.4 & 10.4 & 17.1 & 19.6 & 40.7 & 32.4 \\ 
        &\transg & 42.6 & \textbf{39.4} & 67.0 & 57.5 & \textbf{56.4} & \textbf{53.9} & 51.4 & 52.0 & \textbf{16.3} & \textbf{19.7} & \textbf{31.9} & 36.5 & \textbf{44.3} & \textbf{43.2} \\ 
        &\transn & 32.3 & 30.8 & 62.1 & 52.3 & 54.4 & 51.9 & \textbf{52.2} & \textbf{53.6} & 14.5 & 19.3 & 31.0 & \textbf{37.3} & 41.1 & 40.9 \\ 
        \midrule
        \multirow{7}{*}{Llama-2-13b-Chat} 
        &\nativebasic & 22.7 & 4.9 & 59.5 & 48.4 & 39.9 & 33.7 & 55.2 & 48.2 & 20.7 & 9.6 & 28.4 & 23.8 & 37.7 & 28.1 \\ 
        &\enbasic & 28.7 & 4.4 & 63.9 & 51.6 & 48.2 & 39.8 & \textbf{59.6} & \textbf{56.8} & 20.9 & 17.8 & 31.3 & 30.2 & 42.1 & 33.4 \\ 
        &\nativecot & 26.9 & 4.9 & 59.0 & 49.3 & 38.6 & 33.5 & 56.2 & 47.8 & 17.9 & 7.8 & 28.4 & 22.7 & 37.8 & 27.7 \\ 
        &\encot & 29.5 & 5.5 & 68.2 & 51.0 & 46.2 & 41.8 & 57.8 & 56.6 & 20.5 & 17.3 & 30.7 & 28.0 & 42.1 & 33.4 \\ 
        &\xlt & 32.8 & 6.5 & 68.1 & 52.7 & \textbf{56.9} & \textbf{47.3} & 56.0 & 54.2 & 19.6 & 16.8 & 22.0 & 18.1 & 42.6 & 32.6 \\ 
        &\transg & \textbf{38.4} & \textbf{40.1} & \textbf{77.8} & \textbf{70.4} & 46.1 & 46.1 & 59.2 & 54.6 & \textbf{32.6} & \textbf{37.8} & \textbf{35.1} & \textbf{38.0} & \textbf{48.2} & \textbf{47.8} \\ 
       & \transn & 32.8 & 30.4 & 72.7 & 67.1 & 45.6 & 45.2 & 58.1 & 56.2 & 26.7 & 34.7 & 33.4 & 37.3 & 44.9 & 45.1 \\ 
        \midrule
        \multirow{7}{*}{Llama-2-70B-Chat} 
        &\nativebasic & 35.7 & 5.6 & 64.2 & 48.0 & 43.0 & 36.0 & 53.3 & 50.4 & 28.9 & 10.4 & 30.1 & 26.8 & 42.5 & 29.5 \\ 
        &\enbasic & 42.5 & 7.7 & 70.7 & 52.0 & 52.7 & 41.9 & 61.9 & 52.8 & 25.7 & 21.5 & 30.2 & 35.3 & 47.3 & 35.2 \\ 
        &\nativecot & 35.5 & 5.6 & 65.3 & 46.8 & 41.0 & 35.6 & 56.0 & 49.6 & 25.3 & 9.9 & 26.0 & 25.2 & 41.5 & 28.8 \\ 
        &\encot & 45.6 & 7.0 & 80.7 & 56.3 & 52.7 & 40.9 & 66.5 & 57.0 & 32.7 & 25.7 & 29.8 & 32.0 & 51.3 & 36.5 \\ 
        &\xlt & 49.0 & 8.4 & 76.4 & 54.7 & \textbf{57.3} & 48.4 & 56.6 & 51.6 & 26.5 & 26.7 & 19.3 & 11.5 & 47.5 & 33.6 \\ 
       & \transg & \textbf{55.5} & \textbf{50.0} & \textbf{86.3} & \textbf{79.7} & 55.3 & \textbf{53.0} & 69.4 & \textbf{64.2} & \textbf{38.7} & \textbf{43.1} & \textbf{33.1} & \textbf{36.7} & \textbf{56.4} & \textbf{54.4}\\ 
        &\transn & 46.5 & 39.7 & 83.3 & 75.6 & 53.7 & 51.0 & \textbf{70.5} & 62.4 & 17.8 & 24.7 & 32.4 & 36.2 & 50.7 & 48.3 \\ 
    \bottomrule
    \end{tabular}
    }
    \caption{Average scores of the  high-resource languages and low-resource languages for the six benchmarks in zero-shot setting. 
    The results of PAWS-X and XL-Sum for bloomz-7b1 are not considered since it was already pre-trained on these tasks. The best result for each model is in \textbf{bold}.
    }
    \label{tab:main_NLP_HL}
\end{table*}

%% file: latex/tables/result_mgsm.tex
\begin{table*}[!ht]
    \centering
    \small
    \begin{tabular}{p{2cm}lccccccccccc}
    \toprule       
        \textbf{Model} & \textbf{Prompt type} & \textbf{de} & \textbf{ru} & \textbf{fr} & \textbf{zh} & \textbf{es} & \textbf{ja} & \textbf{sw} & \textbf{th} & \textbf{bn} & \textbf{te} & \textbf{avg} \\
        
        \midrule
        \multirow{7}{*}{ChatGPT}  
       & \nativebasic & 48.8 & 42.8 & 42.8 & 36.0 & 50.0 & 46.0 & 30.8 & 21.6 & 15.6 & 9.6 & 34.4 \\ 
       & \enbasic & 49.2 & 56.0 & 48.4 & 52.4 & 57.2 & 38.4 & 42.0 & 27.2 & 28.8 & 11.2 & 41.1 \\ 
       & \nativecot & 66.0 & 69.6 & 62.4 & 64.4 & 70.0 & 58.0 & 49.2 & 28.4 & 20.8 & 10.0 & 49.9 \\ 
       & \encot & 74.8 & 72.4 & 71.2 & 67.2 & 75.2 & 62.0 & 58.0 & 51.6 & 52.8 & 26.0 & 61.1 \\ 
       & \xlt & 70.8 & 73.6 & 69.6 & 68.8 & 72.8 & 66.8 & 65.6 & 56.8 & 50.8 & 27.2 & 62.3 \\ 
       & \transg & 76.8 & 76.4 & 75.2 & 73.2 & 76.0 & 70.4 & 73.6 & 76.0 & 74.0 & 67.2 & \textbf{73.9} \\ 
       & \transn & 70.0 & 63.2 & 71.2 & 58.4 & 71.6 & 59.2 & 61.2 & 44.4 & 55.6 & 55.2 & 61.0 \\ 
        
        \midrule
        \multirow{7}{*}{bloomz-7b1} 
       & \nativebasic & 1.2 & 1.2 & 2.0 & 2.8 & 1.6 & 0.8 & 0.8 & 0.0 & 1.6 & 1.2 & 1.3 \\ 
        &\enbasic & 2.0 & 1.6 & 2.4 & 2.8 & 1.6 & 1.2 & 2.0 & 1.2 & 3.6 & 2.0 & 2.0 \\ 
      &  \nativecot & 0.0 & 0.4 & 1.2 & 1.6 & 1.6 & 1.2 & 2.4 & 0.4 & 1.2 & 1.6 & 1.2 \\ 
       & \encot & 2.0 & 1.2 & 2.4 & 2.0 & 0.8 & 2.0 & 1.6 & 1.2 & 2.0 & 1.6 & 1.7 \\ 
       & \xlt & 0.8 & 1.2 & 2.0 & 3.2 & 1.6 & 2.4 & 2.0 & 0.8 & 0.8 & 2.4 & 1.7 \\ 
      &  \transg & 3.2 & 2.0 & 2.4 & 2.4 & 2.4 & 2.4 & 2.0 & 3.2 & 2.0 & 4.8 & \textbf{2.7} \\ 
       & \transn & 2.4 & 1.6 & 3.2 & 0.8 & 2.0 & 2.0 & 3.6 & 2.4 & 2.4 & 3.2 & 2.4 \\

        \midrule
        \multirow{7}{*}{\parbox{2cm}{Mistral-7B-Instruct}}
      &  \nativebasic & 7.6 & 14.4 & 12.0 & 19.2 & 30.8 & 8.8 & 4.0 & 4.4 & 6.8 & 4.4 & 11.2 \\ 
       & \enbasic & 38.4 & 36.4 & 31.6 & 28.0 & 42.4 & 25.6 & 7.6 & 9.6 & 16.0 & 2.0 & 23.8 \\ 
      &  \nativecot & 9.6 & 24.0 & 16.8 & 26.8 & 38.4 & 22.8 & 6.0 & 7.6 & 17.2 & 1.2 & 17.0 \\ 
      &  \encot & 39.2 & 42.0 & 36.0 & 33.6 & 42.0 & 30.8 & 8.0 & 21.6 & 18.4 & 4.4 & 27.6 \\ 
      &  \xlt & 43.6 & 51.6 & 45.2 & 38.4 & 45.2 & 34.0 & 10.4 & 23.6 & 19.6 & 6.4 & 31.8 \\ 
      &  \transg & 42.0 & 46.8 & 41.2 & 44.0 & 42.0 & 39.6 & 38.8 & 35.6 & 42.0 & 41.2 & \textbf{41.3} \\ 
       & \transn & 37.6 & 30.0 & 34.0 & 24.8 & 38.0 & 29.6 & 31.6 & 26.4 & 31.2 & 34.0 & 31.7 \\ 

        \midrule
        \multirow{7}{*}{\parbox{2cm}{Llama-2-13b-Chat}}
     &   \nativebasic & 25.2 & 20.0 & 25.6 & 24.4 & 22.0 & 18.8 & 3.6 & 7.2 & 5.2 & 3.6 & 15.6 \\ 
      &  \enbasic & 32.4 & 26.4 & 32.0 & 26.0 & 34.8 & 20.8 & 3.2 & 5.6 & 5.6 & 3.2 & 19.0 \\ 
      &  \nativecot & 29.2 & 23.6 & 29.2 & 27.6 & 28.4 & 23.2 & 2.8 & 7.2 & 6.4 & 3.2 & 18.1 \\ 
      &  \encot & 34.0 & 32.4 & 32.0 & 24.4 & 35.6 & 18.4 & 5.6 & 6.8 & 6.0 & 3.6 & 19.9 \\ 
      &  \xlt & 34.4 & 34.4 & 33.6 & 29.6 & 37.2 & 27.6 & 4.8 & 8.4 & 9.2 & 3.6 & 22.3 \\ 
      &  \transg & 38.0 & 40.4 & 36.8 & 35.6 & 44.8 & 34.8 & 38.4 & 39.2 & 42.8 & 40.0 & \textbf{39.1} \\ 
      &  \transn & 29.6 & 33.2 & 38.8 & 31.2 & 28.0 & 36.0 & 32.0 & 24.8 & 35.6 & 29.2 & 31.8 \\ 
      
        \midrule
        \multirow{7}{*}{\parbox{2cm}{Llama-2-70B-Chat}}
      &  \nativebasic & 34.8 & 28.4 & 38.8 & 38.8 & 41.2 & 32.0 & 4.4 & 8.4 & 7.6 & 2.0 & 23.6 \\ 
      &  \enbasic & 50.4 & 39.2 & 48.0 & 40.0 & 48.0 & 29.6 & 6.0 & 8.8 & 11.6 & 4.4 & 28.6 \\ 
      &  \nativecot & 41.2 & 31.6 & 36.4 & 35.6 & 36.8 & 31.2 & 6.4 & 5.2 & 9.2 & 1.6 & 23.5 \\ 
      &  \encot & 49.6 & 48.0 & 50.0 & 38.0 & 48.4 & 39.6 & 7.6 & 7.2 & 10.4 & 2.8 & 30.2 \\ 
      &  \xlt & 52.0 & 49.6 & 49.6 & 47.2 & 52.0 & 43.6 & 8.0 & 8.0 & 15.6 & 2.0 & 32.8 \\ 
      &  \transg & 56.8 & 56.4 & 54.4 & 54.8 & 56.4 & 54.0 & 51.6 & 46.0 & 51.6 & 50.8 & \textbf{53.3} \\ 
      &  \transn & 49.6 & 43.6 & 49.2 & 41.2 & 50.4 & 45.2 & 43.6 & 32.0 & 42.0 & 41.2 & 43.8 \\ 
      
    \bottomrule
    \end{tabular}
    \caption{Accuracy scores across various languages on the MGSM benchmark.}
    \label{tab:result_mgsm}
\end{table*}

\begin{table*}[!ht]
    \centering
    \small
    \begin{tabular}{p{2cm}lccccccccccc}
    \toprule   
    \textbf{Model} & \textbf{Prompt type} & \textbf{de} & \textbf{ru} & \textbf{fr} & \textbf{zh} & \textbf{es} & \textbf{ja} & \textbf{sw} & \textbf{th} & \textbf{bn} & \textbf{te} & \textbf{avg} \\
    \midrule
    ChatGPT & Trans-ChatGPT & 77.6 & 75.2 & 78.4 & 76.0 & 78.8 & 69.6 & 75.2 & 62.4 & 65.6 & 42.8 & 70.2 \\ 
    Llama-2-70B-Chat & Trans-Llama & 53.6 & 52.0 & 55.2 & 46.8 & 54.4 & 44.0 & 8.8 & 11.2 & 15.2 & 4.8 & 34.6 \\ 
    \bottomrule
    \end{tabular}
    \caption{Accuracy scores across various languages on the MGSM benchmark with self-translate approach.}
    \label{tab:result_mgsm_self-translate}
\end{table*}

%% file: latex/tables/result_xcopa.tex
\begin{table*}[!ht]
    \centering
    \small
    \resizebox{\textwidth}{!}{%
    \begin{tabular}{p{2cm}lcccccccccccc}
    \toprule       
        \textbf{Model} & \textbf{Prompt type} & \textbf{zh} & \textbf{it} & \textbf{vi} & \textbf{tr} & \textbf{id} & \textbf{sw} & \textbf{th} & \textbf{et} & \textbf{ta} & \textbf{ht} & \textbf{qu} & \textbf{avg} \\
        
        \midrule
        \multirow{7}{*}{ChatGPT}  
       & \nativebasic & 88.0 & 91.8 & 74.0 & 81.4 & 85.4 & 77.2 & 65.2 & 85.4 & 49.6 & 63.4 & 50.0 & 72.3 \\ 
       & \enbasic & 90.0 & 89.8 & 85.0 & 86.0 & 87.2 & 78.2 & 75.0 & 81.4 & 58.2 & 65.8 & 54.8 & 76.1 \\ 
       & \nativecot & 87.0 & 92.6 & 72.8 & 80.8 & 83.8 & 75.4 & 66.8 & 84.8 & 48.6 & 63.2 & 55.2 & 72.4 \\ 
       & \encot & 90.4 & 92.2 & 87.0 & 89.6 & 90.2 & 85.6 & 74.8 & 85.8 & 61.4 & 69.2 & 50.2 & 78.6 \\ 
       & \xlt & 89.4 & 91.2 & 87.4 & 88.0 & 88.8 & 82.4 & 76.4 & 91.0 & 60.6 & 76.8 & 50.4 & 79.3 \\ 
       & \transg & 90.8 & 91.6 & 88.4 & 85.8 & 88.8 & 79.4 & 82.6 & 88.2 & 85.6 & 81.6 & 73.2 & \textbf{84.5} \\
        & \transn & 85.6 & 89.2 & 82.4 & 85.8 & 87.0 & 81.4 & 73.8 & 85.4 & 80.6 & 76.2 & 55.6 & 79.7 \\ 
          
          \midrule
          \multirow{7}{*}{bloomz-7b1}
         & \nativebasic & 46.6 & 48.6 & 14.4 & 1.6 & 48.4 & 39.0 & 20.0 & 0.0 & 19.0 & 2.8 & 20.6 & 21.4 \\
        & \enbasic & 78.2 & 55.6 & 68.6 & 50.2 & 62.8 & 56.8 & 49.6 & 50.0 & 71.4 & 50.0 & 50.4 & 56.5 \\
        & \nativecot & 43.4 & 50.0 & 20.2 & 0.6 & 48.6 & 23.0 & 39.2 & 0.0 & 17.6 & 0.0 & 9.4 & 20.9 \\
        & \encot & 67.4 & 53.4 & 63.0 & 50.4 & 57.4 & 51.4 & 49.6 & 49.4 & 64.0 & 49.6 & 50.6 & 53.9 \\
        & \xlt & 63.8 & 49.6 & 62.4 & 45.6 & 64.0 & 49.0 & 51.2 & 46.0 & 52.8 & 48.0 & 36.6 & 50.5 \\
        & \transg & 68.0 & 68.6 & 66.0 & 65.2 & 68.8 & 60.4 & 59.4 & 67.2 & 61.8 & 61.6 & 57.6 & \textbf{63.7} \\
        & \transn & 64.0 & 67.2 & 61.6 & 63.6 & 64.6 & 62.2 & 57.4 & 62.8 & 62.8 & 61.6 & 54.2 & 61.8 \\

        \midrule
        \multirow{7}{*}{\parbox{2cm}{Mistral-7B-Instruct}}
        & \nativebasic & 67.2 & 82.2 & 59.8 & 55.0 & 65.0 & 47.6 & 51.8 & 36.6 & 49.2 & 51.2 & 43.6 & 54.2 \\
        & \enbasic & 48.6 & 43.6 & 35.4 & 30.6 & 43.6 & 37.8 & 39.8 & 28.6 & 35.2 & 29.4 & 25.0 & 34.9 \\
        & \nativecot & 64.0 & 80.4 & 58.6 & 54.6 & 65.4 & 45.4 & 50.0 & 40.0 & 44.2 & 51.2 & 48.2 & 53.8 \\
        & \encot & 55.8 & 52.2 & 44.6 & 43.8 & 52.2 & 39.8 & 46.0 & 32.6 & 29.2 & 39.4 & 28.2 & 40.8 \\
        & \xlt & 82.6 & 81.4 & 70.8 & 66.8 & 77.8 & 47.8 & 64.2 & 53.6 & 52.0 & 56.6 & 44.0 & \textbf{61.5} \\
        & \transg & 69.4 & 64.8 & 66.8 & 61.0 & 68.8 & 52.2 & 62.0 & 60.8 & 59.8 & 52.0 & 43.6 & 59.2 \\
        & \transn & 60.8 & 63.4 & 62.2 & 59.2 & 63.0 & 50.8 & 51.4 & 60.6 & 55.0 & 51.0 & 27.4 & 54.4 \\

        \midrule
        \multirow{7}{*}{\parbox{2cm}{Llama-2-13b-Chat}}
        & \nativebasic & 65.0 & 62.2 & 51.4 & 50.4 & 57.6 & 46.2 & 48.4 & 50.0 & 40.2 & 47.2 & 47.0 & 50.1 \\
        & \enbasic & 61.2 & 74.2 & 56.2 & 52.8 & 62.0 & 52.0 & 50.6 & 50.6 & 50.2 & 46.4 & 48.4 & 54.3 \\
        & \nativecot & 62.8 & 64.6 & 49.6 & 53.8 & 64.8 & 49.8 & 51.8 & 45.4 & 32.6 & 49.8 & 46.6 & 50.9 \\
        & \encot & 67.4 & 71.8 & 65.4 & 51.4 & 68.2 & 48.2 & 49.0 & 46.8 & 48.6 & 50.4 & 45.6 & 54.5 \\
        & \xlt & 65.4 & 72.6 & 66.2 & 57.2 & 70.0 & 47.0 & 49.2 & 50.8 & 50.2 & 50.6 & 46.6 & 56.0 \\
        & \transg & 77.8 & 80.4 & 75.2 & 75.0 & 76.4 & 66.6 & 67.6 & 74.0 & 71.8 & 68.8 & 63.2 & \textbf{71.9} \\
        & \transn & 73.0 & 75.6 & 69.6 & 74.4 & 73.2 & 67.4 & 62.4 & 73.8 & 66.2 & 68.0 & 51.2 & 68.2 \\

        \midrule
        \multirow{7}{*}{\parbox{2cm}{Llama-2-70B-Chat}}
        & \nativebasic & 61.6 & 81.6 & 49.4 & 49.4 & 55.4 & 50.6 & 46.8 & 49.8 & 41.0 & 46.4 & 44.6 & 51.5 \\
        & \enbasic & 74.6 & 79.4 & 58.0 & 53.6 & 63.2 & 48.8 & 50.2 & 49.4 & 50.4 & 49.0 & 51.0 & 55.3 \\
        & \nativecot & 65.8 & 78.0 & 52.2 & 51.8 & 54.8 & 49.2 & 49.2 & 50.2 & 40.0 & 43.2 & 36.2 & 50.5 \\
        & \encot & 80.4 & 88.0 & 73.6 & 65.4 & 77.8 & 53.0 & 50.0 & 56.0 & 48.0 & 49.8 & 50.6 & 61.2 \\
        & \xlt & 79.8 & 82.0 & 67.4 & 64.6 & 74.4 & 49.8 & 51.8 & 55.0 & 47.8 & 46.2 & 48.2 & 58.7 \\
        & \transg & 87.2 & 88.0 & 83.6 & 82.2 & 89.4 & 76.6 & 77.6 & 83.4 & 83.6 & 76.4 & 68.4 & \textbf{80.9} \\
        & \transn & 83.2 & 86.6 & 80.2 & 79.8 & 85.8 & 74.4 & 71.4 & 79.2 & 79.6 & 76.2 & 58.2 & 77.1 \\
    \bottomrule
    \end{tabular}
    }
    \caption{Accuracy scores across various languages on the XCOPA benchmark.}
    \label{tab:result_xcopa}
\end{table*}

%% file: latex/tables/results_xnli.tex
\begin{table*}[!ht]
    \centering
    \small
    \resizebox{\textwidth}{!}{%
    \begin{tabular}{p{2cm}lccccccccccccccc}
    \toprule       
    \textbf{Model} & \textbf{Prompt type} & \textbf{de} & \textbf{ru} & \textbf{fr} & \textbf{zh} & \textbf{es} & \textbf{vi} & \textbf{tr} & \textbf{sw} & \textbf{ar} & \textbf{el} & \textbf{th} & \textbf{bg} & \textbf{hi} & \textbf{ur} & \textbf{avg} \\
        
        \midrule
        \multirow{7}{*}{ChatGPT} 
       & \nativebasic & 59.0 & 58.8 & 60.2 & 54.0 & 60.2 & 49.2 & 51.6 & 51.0 & 50.6 & 58.0 & 39.6 & 54.8 & 42.8 & 40.4 & 52.2 \\ 
        &\enbasic & 68.6 & 58.2 & 67.4 & 62.2 & 68.4 & 63.0 & 65.6 & 65.2 & 62.4 & 64.6 & 56.4 & 65.4 & 55.8 & 59.0 & \textbf{63.0} \\ 
       & \nativecot & 59.4 & 54.2 & 58.0 & 51.8 & 58.6 & 47.6 & 53.0 & 50.8 & 51.2 & 54.6 & 37.2 & 54.4 & 40.2 & 37.4 & 50.6 \\ 
       & \encot & 62.6 & 56.4 & 61.4 & 57.4 & 65.8 & 57.6 & 58.0 & 54.0 & 53.4 & 59.0 & 51.0 & 59.8 & 48.6 & 45.0 & 56.4 \\ 
       & \xlt & 63.0 & 57.8 & 61.4 & 58.4 & 63.4 & 59.8 & 61.4 & 58.0 & 57.8 & 60.4 & 55.0 & 59.6 & 53.2 & 59.2 & 59.2 \\ 
       & \transg & 65.6 & 59.6 & 65.2 & 62.6 & 62.6 & 58.6 & 60.4 & 57.6 & 63.2 & 62.2 & 56.4 & 60.0 & 57.0 & 55.8 & 60.5 \\ 
        &\transn & 63.4 & 62.2 & 61.6 & 57.4 & 62.8 & 55.6 & 59.4 & 58.8 & 62.4 & 63.4 & 54.2 & 61.6 & 52.8 & 53.0 & 59.2 \\ 
          
        \midrule
        \multirow{7}{*}{bloomz-7b1}
       & \nativebasic & 0.4 & 13.4 & 0.2 & 6.6 & 1.4 & 0.0 & 6.8 & 18.2 & 1.6 & 5.2 & 26.6 & 15.4 & 17.8 & 2.8 & 8.3 \\ 
       & \enbasic & 39.8 & 42.8 & 50.8 & 52.4 & 52.2 & 51.4 & 34.2 & 42.4 & 45.6 & 37.2 & 33.8 & 40.4 & 49.2 & 43.0 & 43.9 \\ 
       & \nativecot & 0.4 & 3.0 & 1.2 & 1.2 & 1.2 & 0.2 & 9.0 & 27.2 & 1.6 & 0.8 & 33.4 & 12.4 & 20.0 & 3.8 & 8.2 \\ 
      &  \encot & 36.2 & 35.2 & 37.4 & 42.2 & 37.4 & 37.2 & 33.2 & 34.8 & 36.2 & 33.6 & 33.2 & 34.2 & 37.6 & 34.4 & 35.9 \\ 
       & \xlt & 38.2 & 34.4 & 35.0 & 34.0 & 35.0 & 36.0 & 37.4 & 35.4 & 34.6 & 35.6 & 35.0 & 36.6 & 33.8 & 34.0 & 35.4 \\ 
       & \transg & 45.0 & 43.4 & 44.2 & 44.0 & 45.2 & 44.8 & 43.8 & 44.0 & 44.0 & 44.6 & 44.4 & 44.8 & 43.4 & 44.4 & \textbf{44.3} \\ 
       & \transn & 45.6 & 43.0 & 44.0 & 44.0 & 45.4 & 42.4 & 43.6 & 43.4 & 44.6 & 44.6 & 43.2 & 44.8 & 42.8 & 42.0 & 43.8 \\

        \midrule
        \multirow{7}{*}{\parbox{2cm}{Mistral-7B-Instruct}}
      &  \nativebasic & 50.4 & 55.6 & 59.2 & 46.0 & 59.0 & 33.4 & 38.8 & 33.0 & 34.2 & 34.2 & 39.2 & 46.6 & 37.0 & 33.2 & 42.8 \\ 
      &  \enbasic & 56.4 & 54.6 & 59.8 & 54.0 & 56.8 & 51.4 & 46.8 & 37.6 & 45.8 & 49.4 & 47.0 & 54.4 & 46.4 & 41.8 & 50.2 \\ 
      &  \nativecot & 50.0 & 55.0 & 58.4 & 47.6 & 54.6 & 35.8 & 38.2 & 32.2 & 37.6 & 35.4 & 40.0 & 52.0 & 36.8 & 33.8 & 43.4 \\ 
      &  \encot & 55.0 & 52.2 & 58.0 & 52.4 & 57.0 & 50.4 & 48.0 & 38.0 & 48.6 & 51.2 & 45.8 & 54.2 & 46.8 & 42.0 & 50.0 \\ 
      &  \xlt & 48.2 & 44.6 & 49.6 & 49.4 & 52.4 & 46.0 & 48.0 & 39.0 & 42.2 & 46.4 & 45.4 & 46.6 & 44.0 & 42.6 & 46.0 \\ 
      &  \transg & 58.6 & 54.2 & 59.2 & 52.6 & 59.0 & 55.0 & 54.6 & 53.0 & 56.4 & 58.2 & 48.8 & 56.8 & 52.4 & 50.6 & \textbf{55.0} \\ 
      &  \transn & 57.0 & 52.4 & 55.8 & 50.2 & 58.2 & 53.0 & 54.2 & 49.4 & 53.0 & 56.4 & 47.4 & 55.2 & 50.0 & 49.6 & 53.0 \\ 

        \midrule
        \multirow{7}{*}{\parbox{2cm}{Llama-2-13b-Chat}}
      &  \nativebasic & 41.4 & 40.2 & 44.0 & 38.6 & 42.8 & 32.4 & 34.6 & 31.6 & 32.8 & 34.2 & 34.0 & 37.4 & 31.4 & 33.6 & 36.4 \\ 
      &  \enbasic & 50.2 & 47.4 & 51.6 & 45.0 & 51.8 & 43.0 & 41.8 & 37.8 & 38.8 & 42.6 & 36.4 & 45.0 & 38.4 & 37.8 & 43.4 \\ 
      &  \nativecot & 39.4 & 42.0 & 43.4 & 32.6 & 42.6 & 31.8 & 31.4 & 33.4 & 31.2 & 35.2 & 32.8 & 38.2 & 32.6 & 33.2 & 35.7 \\ 
       & \encot & 45.6 & 46.8 & 48.8 & 44.4 & 46.6 & 44.8 & 41.8 & 38.6 & 43.2 & 43.4 & 38.6 & 46.2 & 42.0 & 40.8 & 43.7 \\ 
       & \xlt & 59.6 & 55.8 & 56.4 & 54.0 & 59.8 & 55.6 & 48.2 & 37.8 & 49.4 & 49.0 & 44.4 & 52.0 & 48.4 & 49.2 & \textbf{51.4} \\ 
      &  \transg & 50.4 & 44.2 & 45.4 & 44.6 & 46.0 & 46.0 & 47.6 & 42.8 & 48.4 & 48.2 & 43.4 & 45.4 & 45.4 & 47.4 & 46.1 \\ 
      &  \transn & 48.6 & 46.6 & 47.0 & 43.2 & 44.6 & 43.6 & 49.0 & 43.2 & 44.0 & 46.0 & 41.6 & 48.6 & 45.6 & 43.4 & 45.4 \\ 

        \midrule
        \multirow{7}{*}{\parbox{2cm}{Llama-2-70B-Chat}}
       & \nativebasic & 44.0 & 42.0 & 45.4 & 42.6 & 45.6 & 38.4 & 38.4 & 32.6 & 35.0 & 37.6 & 33.0 & 41.8 & 34.8 & 34.8 & 39.0 \\ 
        &\enbasic & 53.6 & 54.6 & 57.0 & 49.6 & 55.6 & 46.0 & 42.8 & 32.4 & 50.2 & 46.2 & 38.6 & 52.4 & 37.6 & 34.8 & 46.5 \\ 
       & \nativecot & 40.4 & 42.2 & 45.4 & 38.4 & 41.4 & 38.4 & 36.6 & 32.8 & 35.2 & 37.4 & 32.6 & 41.0 & 33.2 & 36.2 & 37.9 \\ 
       & \encot & 53.6 & 52.8 & 56.4 & 50.4 & 56.8 & 46.0 & 40.6 & 33.4 & 44.6 & 47.8 & 38.2 & 48.2 & 37.6 & 36.6 & 45.9 \\ 
       & \xlt & 56.0 & 59.4 & 59.6 & 55.2 & 61.2 & 52.6 & 51.4 & 36.4 & 44.4 & 55.4 & 44.6 & 57.8 & 51.2 & 45.8 & 52.2 \\ 
       & \transg & 58.8 & 53.4 & 56.8 & 56.4 & 54.8 & 51.8 & 55.4 & 49.6 & 57.2 & 56.4 & 50.2 & 57.4 & 50.8 & 46.6 & \textbf{54.0} \\ 
       & \transn & 56.4 & 52.8 & 54.6 & 49.8 & 58.6 & 50.2 & 53.4 & 51.0 & 52.0 & 56.0 & 48.8 & 52.4 & 49.0 & 45.6 & 52.2 \\

    \bottomrule
    \end{tabular}
    }
    \caption{Accuracy scores across various languages on the XNLI benchmark.}
    \label{tab:result_XNLI}
\end{table*}

%% file: latex/tables/result_paws-x.tex
\begin{table*}[!ht]
    \centering
    \small
    \begin{tabular}{p{2cm}lccccccc}
    \toprule     
    \textbf{Model} & \textbf{Prompt type} & \textbf{de} & \textbf{fr} & \textbf{zh} & \textbf{es} & \textbf{ja} & \textbf{ko} & \textbf{avg} \\
        
        \midrule
        \multirow{7}{*}{ChatGPT} 
        &\nativebasic & 62.0 & 53.6 & 46.6 & 46.6 & 49.0 & 40.6 & 49.7 \\ 
        &\enbasic & 67.6 & 68.0 & 58.8 & 71.4 & 55.8 & 50.4 & 62.0 \\ 
        &\nativecot & 61.8 & 55.0 & 48.6 & 48.8 & 44.0 & 43.4 & 50.3 \\ 
        &\encot & 67.6 & 64.0 & 61.2 & 70.0 & 55.8 & 51.2 & 61.6 \\ 
        &\xlt & 57.4 & 63.8 & 59.8 & 59.2 & 58.2 & 58.2 & 59.4 \\ 
        &\transg & 69.0 & 69.6 & 66.0 & 71.4 & 65.0 & 62.0 & 67.2 \\ 
        &\transn & 67.0 & 70.6 & 68.6 & 70.2 & 65.4 & 63.4 & \textbf{67.5} \\ 
    

        \midrule
        \multirow{7}{*}{\parbox{2cm}{Mistral-7B-Instruct}}
        &\nativebasic & 40.6 & 47.0 & 49.2 & 44.2 & 41.8 & 44.8 & 44.6 \\ 
        &\enbasic & 46.8 & 47.8 & 47.8 & 46.8 & 45.8 & 46.6 & 46.9 \\ 
        &\nativecot & 43.8 & 50.2 & 38.8 & 43.6 & 45.0 & 44.2 & 44.3 \\ 
        &\encot & 46.2 & 47.4 & 47.8 & 47.0 & 44.8 & 46.4 & 46.6 \\ 
        &\xlt & 47.4 & 49.6 & 47.6 & 46.6 & 48.2 & 47.2 & 47.8 \\ 
        &\transg & 51.2 & 49.8 & 54.0 & 49.6 & 52.4 & 52.0 & 51.5 \\ 
        &\transn & 50.6 & 52.8 & 52.4 & 50.8 & 54.2 & 53.6 & \textbf{52.4} \\ 

        \midrule
        \multirow{7}{*}{\parbox{2cm}{Llama-2-13b-Chat}}
        &\nativebasic & 50.8 & 57.2 & 54.0 & 58.0 & 55.8 & 48.2 & 54.0 \\ 
        &\enbasic & 60.2 & 61.0 & 58.6 & 59.8 & 58.2 & 56.8 & \textbf{59.1} \\ 
        &\nativecot & 50.4 & 58.8 & 59.0 & 55.8 & 56.8 & 47.8 & 54.8 \\ 
        &\encot & 59.2 & 55.8 & 58.6 & 59.2 & 56.4 & 56.6 & 57.6 \\ 
        &\xlt & 54.8 & 58.0 & 53.6 & 56.6 & 56.8 & 54.2 & 55.7 \\ 
        &\transg & 56.6 & 62.0 & 59.6 & 61.6 & 56.2 & 54.6 & 58.4 \\ 
        &\transn & 56.2 & 60.0 & 57.4 & 59.4 & 57.6 & 56.2 & 57.8 \\ 

        \midrule
        \multirow{7}{*}{\parbox{2cm}{Llama-2-70B-Chat}}
        &\nativebasic & 53.4 & 49.8 & 55.6 & 61.0 & 46.8 & 50.4 & 52.8 \\ 
        &\enbasic & 62.8 & 66.2 & 58.4 & 67.0 & 55.2 & 52.8 & 60.4 \\ 
        &\nativecot & 53.0 & 53.4 & 53.6 & 65.4 & 54.6 & 49.6 & 54.9 \\ 
        &\encot & 65.0 & 70.8 & 65.0 & 70.2 & 61.6 & 57.0 & 64.9 \\ 
        &\xlt & 57.0 & 61.6 & 57.6 & 57.2 & 49.4 & 51.6 & 55.7 \\ 
        &\transg & 70.6 & 70.6 & 68.0 & 72.2 & 65.6 & 64.2 & 68.5 \\ 
        &\transn & 69.8 & 73.4 & 69.4 & 71.2 & 68.8 & 62.4 & \textbf{69.2} \\ 
        
    \bottomrule
    \end{tabular}
    \caption{Accuracy scores across various languages on the PAWS-X benchmark.}
    \label{tab:result_paws}
\end{table*}

%% file: latex/tables/result_mkqa.tex
\begin{table*}[!ht]
    \centering
    \small
    \begin{tabular}{p{2cm}lcccccccccc}
    \toprule     
    \textbf{Model} & \textbf{Prompt type} & \textbf{de} & \textbf{ru} & \textbf{fr} & \textbf{zh} & \textbf{es} & \textbf{ja} & \textbf{vi} & \textbf{tr} & \textbf{th} & \textbf{avg} \\
        
        \midrule
        \multirow{7}{*}{ChatGPT} 
&        \nativebasic & 44.1 & 30.5 & 46.4 & 31.4 & 40.2 & 20.2 & 33.0 & 39.2 & 33.6 & 35.4 \\ 
&        \enbasic & 36.9 & 30.5 & 43.3 & 28.9 & 44.1 & 43.5 & 34.7 & 32.7 & 34.0 & 36.5 \\ 
&        \nativecot & 43.6 & 22.2 & 46.1 & 30.0 & 38.3 & 33.9 & 34.1 & 38.3 & 31.9 & 35.4 \\ 
&        \encot & 44.6 & 37.4 & 49.7 & 38.5 & 48.0 & 52.4 & 32.6 & 42.0 & 40.5 & 42.9 \\ 
&        \xlt & 36.6 & 31.0 & 39.3 & 31.8 & 44.0 & 43.6 & 37.3 & 37.9 & 37.2 & 37.6 \\ 
&        ransg & 42.0 & 39.2 & 42.7 & 48.6 & 40.8 & 46.4 & 37.8 & 44.2 & 52.3 & \textbf{43.8} \\ 
&        \transn & 39.2 & 34.6 & 26.7 & 31.6 & 29.1 & 45.3 & 41.2 & 41.2 & 45.9 & 37.2 \\ 
          
        \midrule
        \multirow{7}{*}{bloomz-7b1}
&        \nativebasic & 0.6 & 3.0 & 7.6 & 12.1 & 11.2 & 7.5 & 7.6 & 0.0 & 20.9 & 7.8 \\ 
&        \enbasic & 7.5 & 3.7 & 12.3 & 21.4 & 12.2 & 12.3 & 13.3 & 2.1 & 11.0 & 10.6 \\ 
&        \nativecot & 0.2 & 0.9 & 5.9 & 8.6 & 8.3 & 6.0 & 6.7 & 0.0 & 22.2 & 6.5 \\ 
&        \encot & 4.0 & 3.0 & 11.4 & 17.9 & 13.9 & 8.7 & 11.1 & 1.7 & 12.2 & 9.3 \\ 
&        \xlt & 5.7 & 2.8 & 10.2 & 14.8 & 10.1 & 7.1 & 9.6 & 1.4 & 10.4 & 8.0 \\ 
&        \transg & 13.5 & 11.5 & 10.7 & 25.7 & 12.5 & 22.5 & 12.8 & 11.7 & 34.2 & \textbf{17.2} \\ 
&        \transn & 11.7 & 8.7 & 7.2 & 15.2 & 9.3 & 24.5 & 13.1 & 11.2 & 31.3 & 14.7 \\ 

        \midrule
        \multirow{7}{*}{\parbox{2cm}{Mistral-7B-Instruct}}
&        \nativebasic & 8.5 & 5.2 & 8.7 & 7.2 & 9.5 & 7.4 & 8.0 & 2.6 & 13.5 & 7.8 \\ 
&        \enbasic & 7.9 & 5.0 & 7.5 & 5.1 & 8.7 & 6.7 & 6.3 & 5.3 & 10.6 & 7.0 \\ 
&        \nativecot & 9.1 & 5.4 & 7.7 & 8.1 & 8.2 & 7.9 & 7.3 & 2.8 & 13.6 & 7.8 \\ 
&        \encot & 11.2 & 7.8 & 16.0 & 8.4 & 14.9 & 13.1 & 7.9 & 7.6 & 16.4 & 11.5 \\ 
&        \xlt & 9.7 & 7.2 & 10.4 & 8.4 & 10.4 & 10.5 & 9.2 & 6.6 & 14.2 & 9.6 \\ 
&        \transg & 14.6 & 13.8 & 14.9 & 17.7 & 17.0 & 22.5 & 13.4 & 15.1 & 24.4 & \textbf{17.0} \\ 
&        \transn & 13.3 & 12.7 & 10.5 & 14.9 & 11.8 & 24.1 & 13.8 & 13.5 & 25.2 & 15.5 \\ 

        \midrule
        \multirow{7}{*}{\parbox{2cm}{Llama-2-13b-Chat}}
&        \nativebasic & 15.0 & 13.6 & 31.3 & 20.6 & 29.7 & 13.8 & 21.2 & 5.8 & 13.4 & 18.3 \\ 
&        \enbasic & 28.5 & 11.6 & 28.7 & 13.9 & 27.2 & 21.0 & 15.3 & 15.6 & 20.0 & 20.2 \\ 
&        \nativecot & 14.6 & 10.4 & 29.1 & 13.3 & 23.6 & 10.5 & 23.8 & 5.6 & 10.1 & 15.7 \\ 
&        \encot & 28.2 & 12.6 & 31.1 & 11.9 & 28.9 & 15.3 & 15.4 & 18.3 & 16.3 & 19.8 \\ 
&        \xlt & 23.6 & 17.0 & 27.5 & 10.3 & 26.2 & 18.2 & 14.7 & 16.4 & 17.2 & 19.0 \\ 
&        \transg & 31.1 & 29.9 & 34.6 & 35.1 & 31.7 & 35.4 & 30.8 & 31.7 & 43.9 & \textbf{33.8} \\ 
&        \transn & 26.1 & 26.6 & 19.8 & 27.4 & 18.5 & 36.2 & 32.1 & 29.2 & 40.2 & 28.4 \\ 

        \midrule
        \multirow{7}{*}{\parbox{2cm}{Llama-2-70B-Chat}}
&        \nativebasic & 36.7 & 23.8 & 35.2 & 15.9 & 39.3 & 24.7 & 26.7 & 8.6 & 12.1 & 24.8 \\ 
&        \enbasic & 33.2 & 18.1 & 32.9 & 18.8 & 33.7 & 26.6 & 16.3 & 20.7 & 22.3 & 24.7 \\ 
&        \nativecot & 34.8 & 19.5 & 33.9 & 13.1 & 38.5 & 13.1 & 24.1 & 9.2 & 10.6 & 21.9 \\ 
&        \encot & 39.5 & 24.6 & 39.0 & 24.2 & 41.0 & 35.2 & 25.3 & 26.4 & 25.0 & 31.1 \\ 
&        \xlt & 29.8 & 22.4 & 29.6 & 18.0 & 31.3 & 29.5 & 25.0 & 27.3 & 26.1 & 26.6 \\ 
&        \transg & 37.3 & 34.0 & 37.1 & 43.5 & 35.4 & 48.0 & 35.8 & 38.3 & 47.9 & \textbf{39.7} \\ 
&        \transn & 16.7 & 16.4 & 11.9 & 18.5 & 14.9 & 26.8 & 19.7 & 21.8 & 27.6 & 19.4 \\ 
    \bottomrule
    \end{tabular}
    \caption{F1 scores across various languages on the MKQA benchmark.}
    \label{tab:result_mkqa}
\end{table*}

%% file: latex/tables/result_xlsum.tex
\begin{table*}[!ht]
    \centering
    \small
    \begin{tabular}{p{2cm}lcccccc}
    \toprule     
    \textbf{Model} & \textbf{Prompt type} & \textbf{fr} & \textbf{zh} & \textbf{es} & \textbf{vi} & \textbf{tr} & \textbf{avg} \\

        \midrule
        \multirow{7}{*}{ChatGPT} 
&         \nativebasic & 29.2 & 39.3 & 26.9 & 34.4 & 29.9 & 31.9 \\ 
&         \enbasic & 28.9 & 38.8 & 27.8 & 37.9 & 30.0 & \textbf{32.7} \\ 
&         \nativecot & 28.8 & 38.5 & 26.1 & 34.0 & 27.9 & 31.1 \\ 
&         \encot & 25.4 & 35.1 & 26.0 & 33.5 & 28.6 & 29.7 \\ 
&         \xlt & 24.2 & 25.5 & 18.1 & 23.4 & 26.1 & 23.4 \\ 
&         \transg & 27.2 & 36.2 & 26.3 & 32.6 & 28.9 & 30.3 \\ 
&         \transn & 26.4 & 29.7 & 26.1 & 31.5 & 27.7 & 28.3 \\ 
          
        \midrule
        \multirow{7}{*}{bloomz-7b1}
&         \nativebasic & 14.6 & 24.3 & 20.0 & 7.7 & 8.2 & 14.9 \\ 
&         \enbasic & 20.1 & 23.9 & 20.9 & 20.6 & 14.2 & \textbf{19.9} \\ 
&         \nativecot & 18.2 & 25.4 & 24.1 & 1.7 & 8.0 & 15.5 \\ 
&         \encot & 18.0 & 26.1 & 21.6 & 19.3 & 11.3 & 19.3 \\ 
&         \xlt & 12.2 & 19.9 & 19.3 & 14.5 & 5.3 & 14.2 \\ 
&         \transg & 10.0 & 14.2 & 12.1 & 9.0 & 10.7 & 11.2 \\ 
&         \transn & 10.5 & 8.6 & 12.5 & 9.7 & 11.5 & 10.6 \\ 

        \midrule
        \multirow{7}{*}{\parbox{2cm}{Mistral-7B-Instruct}}
&         \nativebasic & 23.0 & 34.0 & 22.3 & 25.8 & 24.4 & 25.9 \\ 
&         \enbasic & 20.9 & 16.5 & 21.5 & 28.0 & 21.1 & 21.6 \\ 
&         \nativecot & 19.7 & 33.6 & 22.1 & 26.4 & 21.1 & 24.6 \\ 
&         \encot & 20.6 & 12.1 & 19.9 & 22.2 & 18.8 & 18.7 \\ 
&         \xlt & 15.4 & 16.5 & 14.7 & 21.7 & 19.6 & 17.6 \\ 
&         \transg & 26.8 & 34.9 & 26.4 & 39.5 & 36.5 & \textbf{32.8} \\ 
&         \transn & 26.8 & 30.0 & 26.6 & 40.6 & 37.3 & 32.2 \\ 

        \midrule
        \multirow{7}{*}{\parbox{2cm}{Llama-2-13b-Chat}}
&         \nativebasic & 27.7 & 21.9 & 25.3 & 38.8 & 23.8 & 27.5 \\ 
&         \enbasic & 25.7 & 38.2 & 23.6 & 37.7 & 30.2 & 31.1 \\ 
&         \nativecot & 27.9 & 29.0 & 24.8 & 31.8 & 22.7 & 27.2 \\ 
&         \encot & 24.0 & 39.4 & 23.1 & 36.4 & 28.0 & 30.2 \\ 
&         \xlt & 24.2 & 17.7 & 22.4 & 23.6 & 18.1 & 21.2 \\ 
&         \transg & 28.0 & 42.9 & 27.9 & 41.6 & 38.0 & \textbf{35.7} \\ 
&         \transn & 27.5 & 37.5 & 26.9 & 41.6 & 37.3 & 34.2 \\ 

        \midrule
        \multirow{7}{*}{\parbox{2cm}{Llama-2-70B-Chat}}
&         \nativebasic & 28.8 & 34.5 & 27.3 & 29.7 & 26.8 & 29.4 \\ 
&         \enbasic & 29.0 & 31.8 & 24.3 & 35.7 & 35.3 & 31.2 \\ 
&         \nativecot & 25.3 & 29.5 & 26.7 & 22.4 & 25.2 & 25.8 \\ 
&         \encot & 27.0 & 35.2 & 22.1 & 34.8 & 32.0 & 30.2 \\ 
&         \xlt & 18.1 & 29.7 & 15.2 & 14.2 & 11.5 & 17.7 \\ 
&         \transg & 26.8 & 39.7 & 27.1 & 38.7 & 36.7 & \textbf{33.8} \\ 
&         \transn & 26.6 & 37.5 & 26.3 & 39.0 & 36.2 & 33.1 \\  
    \bottomrule
    \end{tabular}
    \caption{ROUGE-1 scores across various languages on the XL-sum benchmark.}
    \label{tab:result_xlsum}
\end{table*}